\pdfoutput=1
\documentclass{article}

    \PassOptionsToPackage{numbers, compress}{natbib}

\usepackage[preprint]{neurips_2024} 

\usepackage[utf8]{inputenc} 
\usepackage[T1]{fontenc}    
\usepackage{hyperref}       
\usepackage{url}            
\usepackage{booktabs}       
\usepackage{amsfonts}       
\usepackage{nicefrac}       
\usepackage{microtype}      
\usepackage{xcolor}         

\usepackage{amsmath}
\usepackage{hyperref}
\usepackage{multirow} 
\usepackage{makecell} 
\usepackage{wrapfig}

\title{Cognitively Inspired Energy-Based World Models}
\newcommand{\pa}{EBWM}
\newcommand{\parch}{EBT} 
\newcommand{\baseline}{TAMs} 

\newcommand{\ebma}{Regressive Energy Based Modeling} 
\newcommand{\el}{Representation-Induced Labeling}

\usepackage{pifont}
\newcommand{\cmark}{\textcolor{green}{\ding{51}}}%
\newcommand{\xmark}{\textcolor{red}{\ding{55}}}%

\usepackage[draft,textsize=footnotesize,textwidth=15mm]{todonotes}

\author{\textbf{Alexi Gladstone$^\ddagger$\thanks{Correspondence to Alexi Gladstone: \texttt{alexigladstone@gmail.com}, \href{https://alexiglad.github.io/}{https://alexiglad.github.io/}.} , Ganesh Nanduru$^\ddagger$, Md Mofijul Islam$^\ddagger$$^\S$\thanks{Work does not relate to position at Amazon.}}\\
\textbf{Aman Chadha$^\bullet$$^\S$$^\dagger$, Jundong Li$^\ddagger$, Tariq Iqbal$^\ddagger$} \\
$^\ddagger$University of Virginia, $^\bullet$Stanford University, $^\S$Amazon GenAI
\vspace{-0.3in}
}

\begin{document}

\maketitle

\begin{abstract}
\label{sec:abstract}
One of the predominant methods for training world models is autoregressive prediction in the output space of the next element of a sequence. In Natural Language Processing (NLP), this takes the form of Large Language Models (LLMs) predicting the next token; in Computer Vision (CV), this takes the form of autoregressive models predicting the next frame/token/pixel. 
However, this approach differs from human cognition in several respects.
First, human predictions about the future actively influence internal cognitive processes. 
Second, humans naturally evaluate the plausibility of predictions regarding future states. 
Based on this capability, and third, by assessing when predictions are sufficient, humans allocate a dynamic amount of time to make a prediction. This adaptive process is analogous to System~2 thinking in psychology.
All these capabilities are fundamental to the success of humans at high-level reasoning and planning.
Therefore, to address the limitations of traditional autoregressive models lacking these human-like capabilities, we introduce Energy-Based World Models (EBWM). EBWM involves training an Energy-Based Model (EBM) to predict the compatibility of a given context and a predicted future state. In doing so, EBWM enables models to achieve all three facets of human cognition described. 
Moreover, we developed a variant of the traditional autoregressive transformer tailored for Energy-Based models, termed the Energy-Based Transformer (EBT).
Our results demonstrate that EBWM scales better with data and GPU Hours than traditional autoregressive transformers in CV, and that EBWM offers promising early scaling in NLP. Consequently, this approach offers an exciting path toward training future models capable of System 2 thinking and intelligently searching across state spaces. 

\end{abstract}    
\section{Introduction}
\label{sec:intro}

\begin{figure*}[!ht]
    \centering
    \begin{tabular}{cccc}
        \includegraphics[width=0.18\columnwidth]{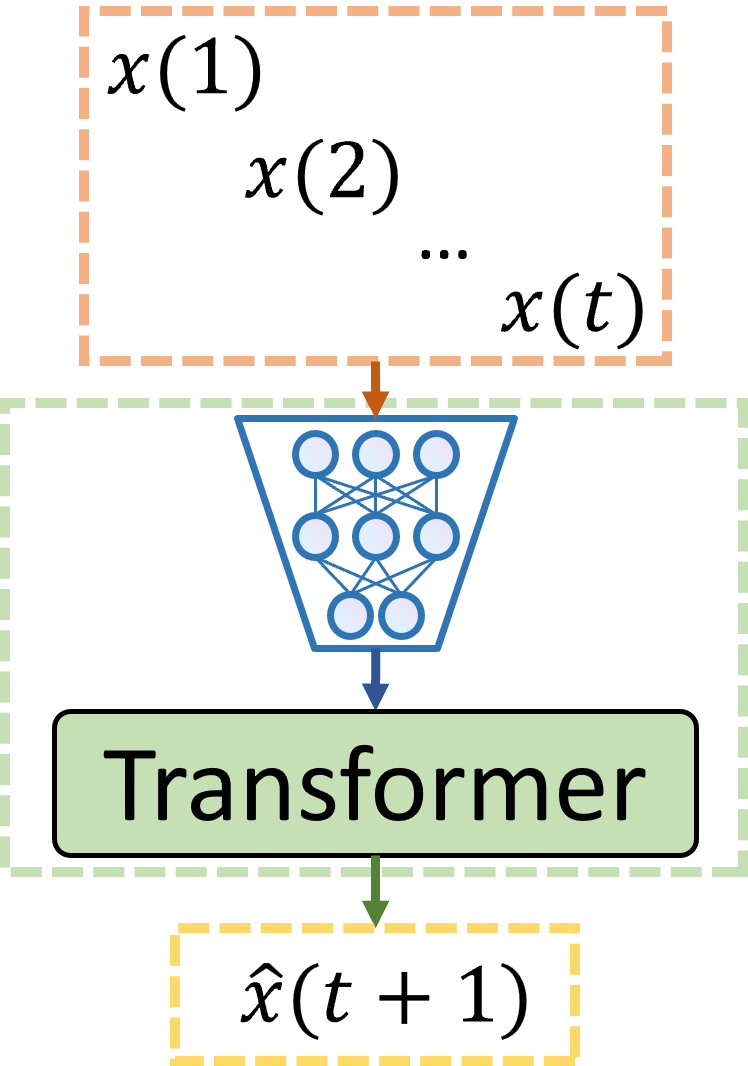} &
        \includegraphics[width=0.2\columnwidth]{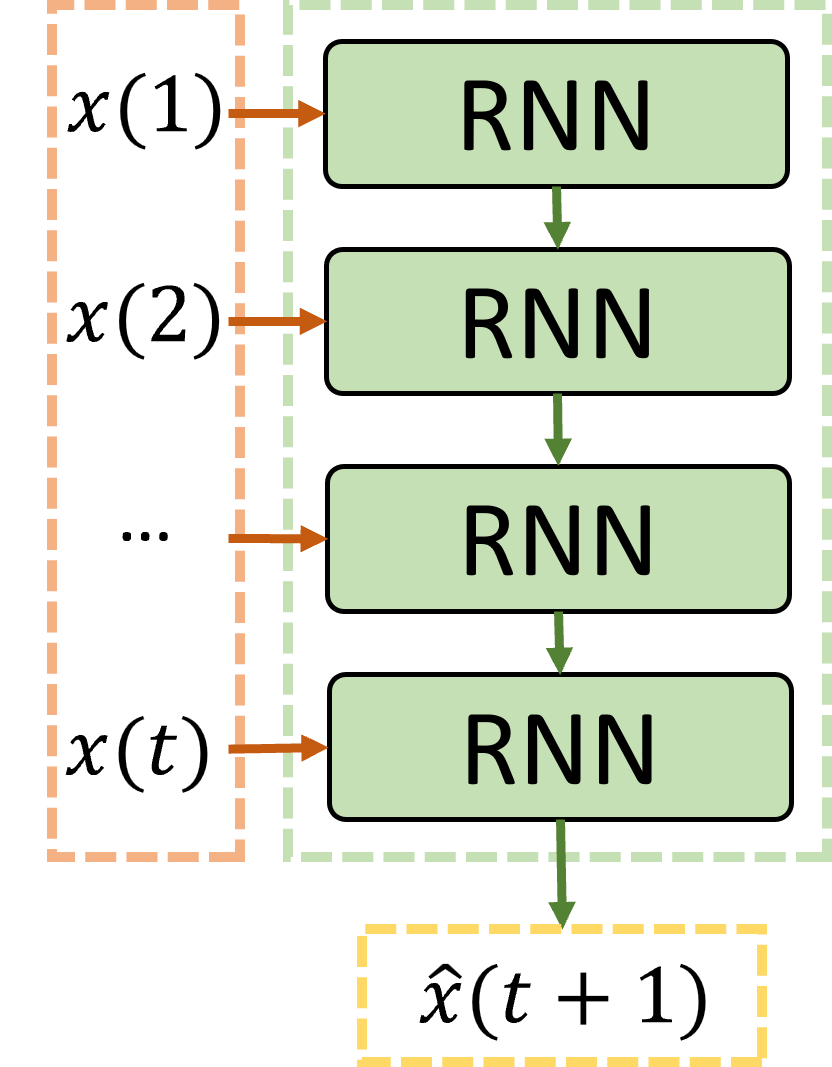} &
        \includegraphics[width=0.24\columnwidth]{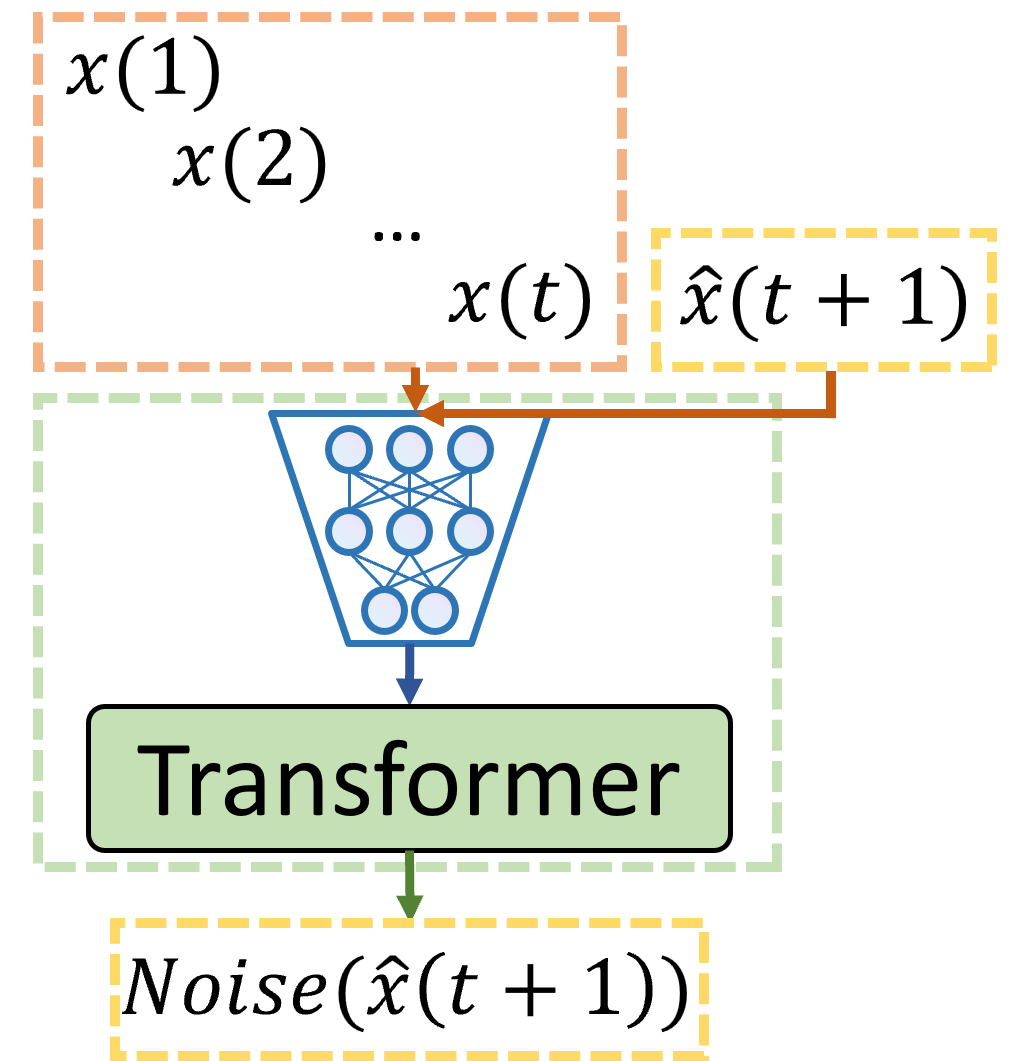} &
        \includegraphics[width=0.24\columnwidth]{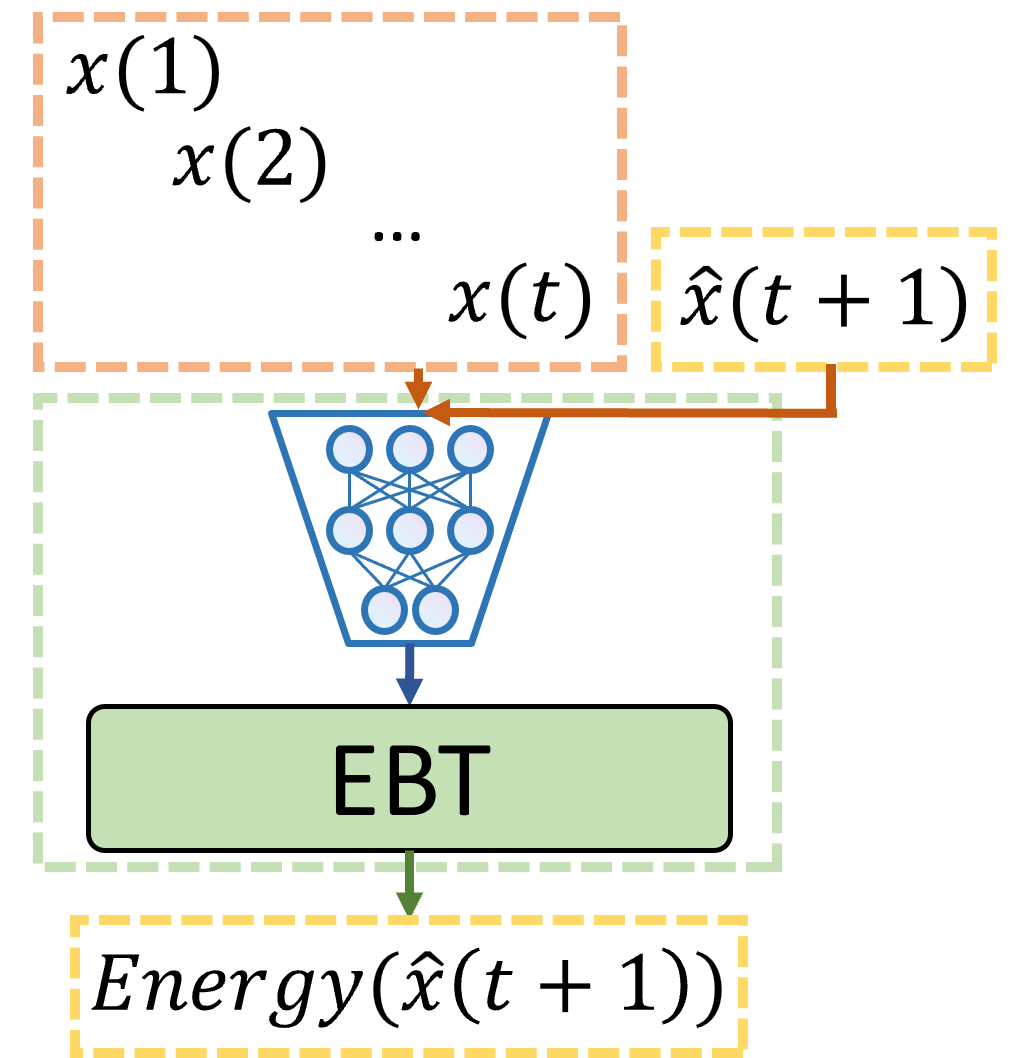} \\
        (a) \label{fig:transformer_arch} AR Trans. &
        (b) \label{fig:rnn_arch} RNN &
        (c) \label{fig:diffusion_arch} Diffusion + Trans. & \multirow{2}{*}{(d) {\pa}}  \\ \cline{1-3}
        \multicolumn{3}{c}{Existing Autoregressive Approaches} & \\
    \end{tabular}
    \caption{Comparison of traditional autoregressive approaches and {\pa}. (a) Autoregressive (AR) Transformer is the most common, and what we compare scaling results to in this paper, with (b) RNN becoming more popular within the past year \cite{gu2023mamba, peng2023rwkv}. (c) is the most similar to {\pa}, being able to allocate a dynamic amount of computation during inference, but predicts the noise rather than the energy \cite{höppe2022diffusion}. Several architectures are interchangeable, such as RetNet for Transformers \cite{sun2023retentive}, or Mamba/LSTM \cite{gu2023mamba, hochreiter1997long} for RNNs. {\pa}, being an Energy-Based Model (EBM), has all four cognitive capabilities described. Trans. stands for Transformer.} 
    \label{fig:model_comparison}
\end{figure*}

Self-Supervised Learning (SSL) has established itself as a powerful method for training large foundation models in Computer Vision (CV) \cite{bardes2023mcjepa, oquab2023dinov2, he2021masked, caron2021emerging, he2020momentum, chen2020simple}, Natural Language Processing (NLP) \cite{openai2023gpt4, zhao2023survey, brown2020language, alayrac2022flamingo, devlin2019bert, touvron2023llama}, and audio/speech processing \cite{benita2023diffar, hsu2023parallel, borsos2023audiolm}. Owing to several of these foundation models having a general understanding of a variety of concepts, these models have found numerous use cases. For example, Large Language Models (LLMs) have been used to augment productivity \cite{noy2023experimental, Spataro_2023}, video generation models have been used to generate realistic videos \cite{stablevideodiffusion}, and audio/speech models have been used for conversational AI \cite{HeySiri, OpenAI_2023}.

Within SSL, one common training approach for achieving such general representations has been autoregressive models predicting the next element of a sequence in the output space. In NLP, this takes the form of predicting the next token \cite{radford2019language}, in CV, models are trained to predict the next frame \cite{mahjourian2017geometry, zhou2020deep}, next frame's feature \cite{oord2018representation} or a discretized token (i.e. via Vector Quantization  \cite{islam2023eqa, van2017neural, bai2023sequential, hu2023gaia1, islam2022representation, yasar2023vader}), and in Audio Processing, a commonly used approach involves predicting discretized speech units \cite{lee2021direct, hsu2023parallel, valle2020flowtron} or continuous waveforms \cite{benita2023diffar}. It has even been shown that some approaches can be generalized across modalities \cite{hawthorne2022generalpurpose}. These autoregressive models, which predict the next element of a sequence in the output space, can be broadly classified as world models (more on this definition is in Section \ref{sec:world_models}), and more specifically, as Traditional Autoregressive Models ({\baseline}). These are shown in Fig.~\ref{fig:transformer_arch}(a) and Fig.~\ref{fig:transformer_arch}(b).

\phantomsection
\textbf{World models compared to human cognition:}
\label{sec:cognitive_facets}
World models have achieved remarkable feats in recent years, scoring higher than the average human on the SAT and LSAT \cite{openai2023gpt4} as well as being able to generate synthetic photo-realistic videos of hypothetical scenarios \cite{OpenAI_sora}. Yet, these models still struggle to perform seemingly simple and innate human capabilities, such as reasoning, planning, or thinking about problems over an extended horizon of time \cite{valmeekam2022large, zhao2024exploring, li2024survey, liu2023llm, mondorf2024beyond}. This phenomenon is broadly known as Moravec's Paradox \cite{agrawal2010study}. Along similar lines, recent work \cite{west2023generative} has shown that many generative AI models are better at generating than they are at understanding, achieving expert performance on tasks such as image generation but failing to answer simple questions about images that a preschooler could answer. This also contrasts from humans, who generally develop the skill of understanding before generating \cite{west2023generative, alexander2003development}. Thus, it's possible a significant portion of the disparity between modern AI and human cognition is caused by fundamental differences in the way {\baseline} work as compared to the human brain. Achieving human-like cognition in models could enhance their ability to reason, plan, and perform System 2 thinking at the same level as humans. As such, in this work, we design an architecture to achieve the following four core cognitive capabilities believed to be fundamental to high-level reasoning and planning in humans:

\phantomsection
\textbf{Facet (1): Predictions shape internal state:}
\label{item:predictions_shape_state} 
It is widely supported that when humans predict the future, these thoughts naturally shape our brains' internal state \cite{bubic2010prediction, Tomasi2015Dissecting, Schacter2007Remembering, huang2011predictive}. However, {\baseline}, having predictions made in the output space, do not have their internal state shaped by a given prediction (see Fig.~\ref{fig:model_comparison}). \footnote{While one could argue that re-incorporating these predictions within the context of models can alter the model's internal state, it is crucial to note that in the realm of {\baseline}, such predictions are treated as ground truth (through teacher forcing during training). Consequently, these models are incapable of discerning between actual observations and their own predictions.}

\phantomsection
\textbf{Facet (2): Evaluation of predictions:}
\label{item:prediction_evaluation}
There is strong evidence supporting the idea that humans naturally evaluate the plausibility of predictions \cite{Nayebi2023Neural, Connell2006A, Brown2012The}. {\baseline}, making predictions of future states in the output space, cannot evaluate the strength or plausibility of predictions. 

\phantomsection
\textbf{Facet (3): Dynamic allocation of resources:}
\label{item:dynamic_computation}
The idea that humans naturally dedicate various amounts of time towards making predictions or reasoning is widely supported in psychology and neuroscience \cite{kahneman2011thinking, fuster1991prefrontal, rougier2005prefrontal}. As the difficulty of tasks humans face varies widely, the ability to adjust the magnitude of computational resources allocated towards a task is fundamental to success. 

\phantomsection
\textbf{Facet (4): Modeling uncertainty in continuous state spaces:}
\label{item:prediction_uncertainty}
LLMs can naturally model uncertainty through the allocation of probability mass across token classes \cite{tomani2024uncertainty}. In the context of continuous state spaces, such as in CV, without the usage of discretization schemes this is not possible with {\baseline}. This capability is particularly important in situations where uncertainty is essential, and is a natural capability of humans \cite{Peters2017Uncertainty, Vilares2012Differential, Sarinopoulos2010Uncertainty}.

\begin{table*}[t]
	\centering
 \caption{All four architectures shown in Fig.~\ref{fig:model_comparison} and whether or not they have the four cognitive facets described in Section \ref{sec:cognitive_facets}. The transformer architecture refers to standard autoregressive transformers. We classify Traditional Transformers and RNNs as {\baseline}. *Diffusion models are not generally deemed architectures but rather an approach. Despite being less common than autoregressive approaches, we wanted to include them since they have more facets than {\baseline}.}
	\begin{tabular}{lcccc}
            \toprule
            \multirow{2}{*}{\textbf{Architecture}} & \multirow{2}{*}{\makecell{\textbf{Predictions affect} \\ \textbf{internal state \ref{item:predictions_shape_state}}}} & \multirow{2}{*}{\makecell{\textbf{Evaluation of} \\ \textbf{predictions \ref{item:prediction_evaluation}}}} & \multirow{2}{*}{\makecell{\textbf{Dynamic compute} \\ \textbf{allocation \ref{item:dynamic_computation}}}} & \multirow{2}{*}{\makecell{\textbf{Uncertainty} \\ \textbf{modeling \ref{item:prediction_uncertainty}}}} \\
            & & & & \\
            \toprule
            Transformers & \xmark & \xmark & \xmark & \xmark \\
            RNN Variants & \xmark & \xmark & \xmark & \xmark \\
            Diffusion* & \cmark & \xmark & \cmark & \xmark \\
            {\pa} & \cmark & \cmark & \cmark & \cmark \\
            \bottomrule
        \end{tabular}
 \label{tab:cognitive_facets}
\end{table*}

To achieve all four cognitive facets described, we approach the problem of world modeling as \textit{making future state predictions in the input space} and predicting the \textit{energy/compatibility} of these future state predictions with the current context through the usage of an Energy-Based Model (EBM). Making predictions in the input space achieves cognitive facets (1) and (3), while predicting the compatibility of future state predictions with the current context achieves cognitive facets (2) and (4) (shown in Table \ref{tab:cognitive_facets}). The architecture of {\pa} is shown in Fig~\ref{fig:model_comparison}(d).

Additionally, as EBMs have struggled to compete with models using modern architectures, we design a domain-agnostic EBM transformer architecture named the Energy-Based Transformer ({\parch}). We will move the whole EBM paradigm forward by releasing the implementation of {\parch}, which allows for the parallelization of multiple predictions at once, similar to traditional transformers \cite{vaswani2017attention}. Our experiments demonstrate the scalability of {\parch}, achieving better scaling in terms of data and GPU hours in CV when compared to {\baseline} on different datasets.

The key contributions of our work can be summarized as follows:
\begin{itemize}
    \item We propose a new architecture that is inspired by four core facets of human cognition for training world models, {\pa}, and lay the ground work for the techniques necessary to train and implement {\pa} across domains. We do extensive analyses regarding different design choices to further support {\pa}'s design.
    \item We design and experiment with a domain-agnostic variant of the transformer architecture, {\parch}, specifically for EBMs.
    \item We implement {\pa} for training world models in CV and NLP. Our findings indicate that {\pa} scales similarly to traditional autoregressive transformers, despite being qualitatively different in allowing for the four facets of human cognition described.
\end{itemize}

\section{Background: World Model primer}

\label{sec:world_models}
Autoregressive models predicting the next element of a sequence can be broadly classified as a specific type of world model. Below, we justify this through a precise definition, while giving some additional background to clarify synonymic terms.

World models have been broadly referred to as ``internal models of how the world works'' \cite{lecun2022path} and ``predictive model of the future that learns a general representation of the world'' \cite{hu2023gaia1}. More precisely, we define world models most broadly as models that given a state, a previous estimate of the state, a latent variable, and an action, predict the next state \cite{lecunworldmodeldef, hu2023gaia1, ha2018world, lecun2022path}. This idea can be formalized as the following (\cite{lecunworldmodeldef}):
\begin{align}
s(t+1) = F(s(t), x(t), a(t), z(t)) \label{eq:world_model_general},
\end{align}

where $x(t)$ is an observation, $s(t)$ is a previous estimate of the world, $a(t)$ is an action, $z(t)$ is a latent variable, and $F$ is the function utilized \cite{lecunworldmodeldef}. The previous estimates of the world are often encapsulated within past observations. Additionally, most world models used in reinforcement learning are not conditioned on a latent variable, and therefore take the following form: 

\begin{align}
s(t+1) = F(x(1), x(2), ..., x(t), a(t))     \label{eq:world_model_rl}.
\end{align}

Therefore, traditional autoregressive models that have been used in NLP (LLMs), vision (video generation models), and audio processing (speech generation models), can be seen as a specific type of world model where the state is defined by the observations and no action or latent variable is being conditioned on \cite{liu2024world, lecunworldmodeldef}. These types of world models simplify to the following mathematically, where $(x(1), ... , x(t))$ is often referred to as the context:

\begin{align}
x(t+1) = F(x(1), x(2), ..., x(t))       \label{eq:world_model_pc}
\end{align}

Furthermore, it's worth noting that the idea of world models has a backing in both robotics and in neuroscience. World models are analogous to Model Predictive Control in robotics \cite{morari1999model, lecun2022path}. Similarly, world models, when the observations are sensory information and the state is approximated using sensory information, are comparable to predictive coding \cite{huang2011predictive} from neuroscience. The primary difference lies in the application of these terms \cite{taniguchi2023world}. In this work we use the term world model to broadly refer to the most general world model definition, Model Predictive Control, and predictive coding.  
\section{Related Work}
\subsection{Traditional autoregressive world models}

Several world models across domains have been proposed in the literature, including LLMs \cite{touvron2023llama, jiang2023mistral, team2023gemini, liu2024world}, video world models \cite{bardes2023v, hoppe2022diffusion, weissenborn2019scaling}, and autoregressive audio processing world models \cite{borsos2023audiolm, benita2023diffar, lu2023unifiedio}. Similarly, in Reinforcement Learning, world models conditioned on an action are common \cite{chen2022transdreamer, hafner2019dream, yang2023learning, hu2023gaia1}.

\textbf{Transformers and Variants:} Several world models \cite{radford2019language, touvron2023llama, latif2023transformers} have used the transformer architecture \cite{vaswani2017attention}, which has become ubiquitous in sequence modeling due to their efficient usage of computation. In the context of traditional autoregressive transformers making predictions of the next state in the output space, as in Fig.~\ref{fig:model_comparison}(a), these models also lack several of the cognitive facets discussed in Section \ref{sec:cognitive_facets} (Table \ref{tab:cognitive_facets}). This includes the ability to leverage a dynamic amount of computation for making predictions or the ability to evaluate the plausibility of predictions.

\textbf{RNNs and Variants:} Recently, several RNN variants (shown in Fig.~\ref{fig:model_comparison}(b)) have emerged to reduce memory bottlenecks and achieve faster inference \cite{gu2023mamba, peng2023rwkv}. These approaches have scaled similarly to transformers in autoregressive sequence modeling, and achieve better memory efficiency and reduced latency. However, traditional RNNs making predictions of the next state in the output space also lack several of the cognitive facets discussed in Section \ref{sec:cognitive_facets} (see Table \ref{tab:cognitive_facets}).


\subsection{Autoregressive world models with dynamic computation}

\subsubsection{Diffusion}
Several existing works have attempted to enable pre-trained autoregressive models to leverage extra computation during inference to make high level decisions or solve challenging problems.
The most common instance of this is diffusion models (Fig.~\ref{fig:model_comparison}(c)), where using multiple forward passes for generating a prediction is a core aspect of both training and inference \cite{höppe2022diffusion, rombach2022high}. Although this idea was originally developed in CV, it has recently been popularized in audio processing \cite{benita2023diffar} as well as NLP \cite{zou2023survey}. These models also lack two of the cognitive facets discussed in Section \ref{sec:cognitive_facets} (Table \ref{tab:cognitive_facets}), through not having the ability to evaluate predictions or model uncertainty in continuous state spaces.


\subsubsection{Dynamic computation in LLMs}
The ability to leverage a dynamic amount of computation is closely mirrored in chain-of-thought prompting, where an LLM is prompted to elicit its thought process before giving an answer \cite{wei2022chain}. This gives an LLM additional computational depth based off the number of tokens decoded before making a prediction, and as a result significantly improves performance in many cases. Similarly, a recently developed solution via the concept of a thinking token \cite{goyal2023think} enabled LLMs to leverage multiple forward passes in predicting the next token. However, both these approaches do not achieve all four cognitive facets described in Section \ref{sec:cognitive_facets} (see Section \ref{sec:counterarguments} for more details).


\subsection{Energy-Based Models}

One contribution of this work was the design of a custom architecture for EBM's called the Energy-Based Transformer ({\parch}). Somewhat similar is the work of the Energy Transformer \cite{hoover2024energy}. Despite strong similarity in the names of these architectures, however, they are very different--with the primary similarity in architectures being the usage of attention mechanisms as well as a global energy function. The existing work integrated ideas from Modern Hopfield Networks, including RNNs, whereas in our work the architecture is non-recurrent and does not use associative memories. Additionally, {\parch} differs with its focus on autogressive modeling, which this previous work did not experiment with.

The most similar works to ours involve autoregressive Energy-Based Models, including E-ARM \cite{wang2022your}, EBR \cite{bhattacharyya2020energy}, and Residual EBMs \cite{bakhtin2021residual}. E-ARM involves adding an objective to the learning process to turn a traditional autoregressive model into an EBM, and as such does not achieve three of the cognitive characteristics discussed in Section \ref{sec:cognitive_facets}. EBR and Residual EBMs involve the training of an EBM on top of an already pre-trained autoregressive language model. Both works, however, leverage a contrastive objective, which suffers from the curse of dimensionality. {\pa} differs from these works through leveraging a reconstruction objective (more details in Section \ref{sec:energy}), being domain agnostic, and by not requiring a pre-trained model to work on top of ({\pa} is a standalone solution). We also develop and leverage the {\parch} to achieve better scaling--previous works did not use such modern architectures.


\section{Energy-Based World Models ({\pa}) Intuition}

\label{sec:intuition}
Energy-Based Models (EBMs) are a class of models that associate a scalar energy value with each configuration of input variables, producing lower values for highly compatible inputs and higher values for less compatible inputs \cite{lecun2022path}. {\pa} leverages this, and is trained to predict \textit{how compatible} a given context and predicted future state are, where high energy corresponds to low compatibility and low energy corresponds to high compatibility.

\textbf{Intelligent search:}
Recent work has revealed a characteristic of world models, deemed ``The Generative AI Paradox,'' where these models have achieved superhuman generative capabilities, being able to outperform even the best humans, but paradoxically lack the ability to discriminate or ``understand'' states at the level of an average human \cite{west2023generative} or reason at the level of an infant \cite{stojnic2023commonsense}. This is problematic as the ability to discriminate is essential in determining what states are plausible and desired when reasoning and planning. For example, in the realm of NLP, researchers have explored approaches such as tree search for generating several plausible answers \cite{Team_2023, yao2023tree}. One limitation of this approach that has often been stated is that LLMs know how to generate the right answer, but they do not know how to \textit{choose} this answer \cite{Meta_2024}. This has made tree search extremely computationally expensive, as thousands or even millions of samples need to be generated to achieve optimal performance \cite{Team_2023}. Therefore, world models with the ability to directly evaluate the plausibility of a proposed state (\ref{item:prediction_evaluation}) offer a promising path in solving this challenge of searching the state space intelligently. {\pa}, being an EBM, can use its scalar energy value to directly evaluate a state and MCMC (more details on MCMC can be found in Section \ref{sec:mcmc}) to directly improve upon a state--thereby offering an opportunity to both select states as well as efficiently search the state space. This scalar energy value also offers benefits in determining when to finish generating, as a cutoff threshold for the energy value can be used to determine when the model has achieved sufficient certainty about a prediction.

\textbf{System 2 thinking:}
Consider the task of a person making the most important business decision of their life. Rather than making this decision based on a single fleeting thought, analogous to System 1 thinking from psychology, it's highly probable this person would use System 2 thinking \cite{kahneman2011thinking}, and ponder over a long period of time. This ability for humans to think over an indefinite period of time is believed to be essential for reasoning \cite{kahneman2011thinking, evans2003two, alter2007overcoming}. {\baseline}, using a single forward pass to make a prediction of the next state, cannot use a dynamic amount of computation for a prediction of the future state, making it challenging to reason (see Section \ref{sec:counterarguments} for common counterarguments to this perspective). Diffusion models, in allowing for a dynamic amount of inference cycles when denoising, do allow for dynamic resource allocation (Section \ref{item:dynamic_computation}), but are still challenged by the fact that they cannot evaluate the plausibility of future state predictions \ref{item:prediction_evaluation}. {\pa}, achieving all four cognitive facets described, can use a dynamic amount of computation and evaluate its predictions, offering a promising path towards System 2 thinking.

\textbf{Modeling Uncertainty:}
Consider the case of a vision world model being rolled out to decide whether the left or right fork in a road will be taken. With {\baseline}, only a single next frame/token/embedding can be predicted as predictions are made in the output space as shown in Fig.~\ref{fig:model_comparison}. Consequently, {\baseline} would not be able to model the left and right fork in this situation. In cases like these, the modeling of different potential futures is important, and one reason LLMs have succeeded (as states are discrete in LLMs and take the form of tokens). {\pa}, being an EBM, can naturally evaluate the energy of different predicted futures, allowing for the modeling of uncertainty.

\section{Energy-Based World Models ({\pa}) Approach}

\subsection{Energy-based Models}
\label{sec:energy}

In traditional approaches to training Energy-Based Models (EBMs), two primary methods have been prevalent: contrastive methods and regularized methods. Contrastive learning approaches \cite{lecun2022path}, such as the most commonly used EBM training approach of contrastive divergence \cite{hinton2002training, du2020improved}, succumb to the curse of dimensionality. Although regularized methods avoid this flaw, they struggle with imposing non-restrictive regularization or inductive biases such as bottlenecks \cite{lecun2022path}. 

A recent study demonstrated a regularized approach with weak restrictions \cite{wang2023energy}, incorporating a reconstruction objective. This approach involved training an EBM to create representations such that minimizing the energy predicts a denoised image. Therefore, rather than using a contrastive loss, this approach involved a reconstruction loss calculated in the input space. For continuous state spaces, such as visual signals, this takes the form of a SmoothL1 loss calculated between predicted and ground truth embeddings with $\beta = 1$; and for discrete state spaces, such as in NLP, categorical cross-entropy is used. To our knowledge, this loss has not been used within existing autoregressive EBMs, and aligns with the ideals of training a world model to predict the next state conditioned on all previous observations. We find this simple objective consistently achieves the most stable runs and best loss when compared to other losses (more details on other losses in the supplementary section \ref{sec:supp_losses}). As stated in \cite{wang2023energy}, this can intuitively be seen as ``folding the encoder-decoder architecture into the forward and backward passes of a model.'' Detailed pseudocode for the implementation of this approach is in \cite{wang2023energy} Section A.3.

\begin{figure}
    \centering
    \includegraphics[width=.85\textwidth]{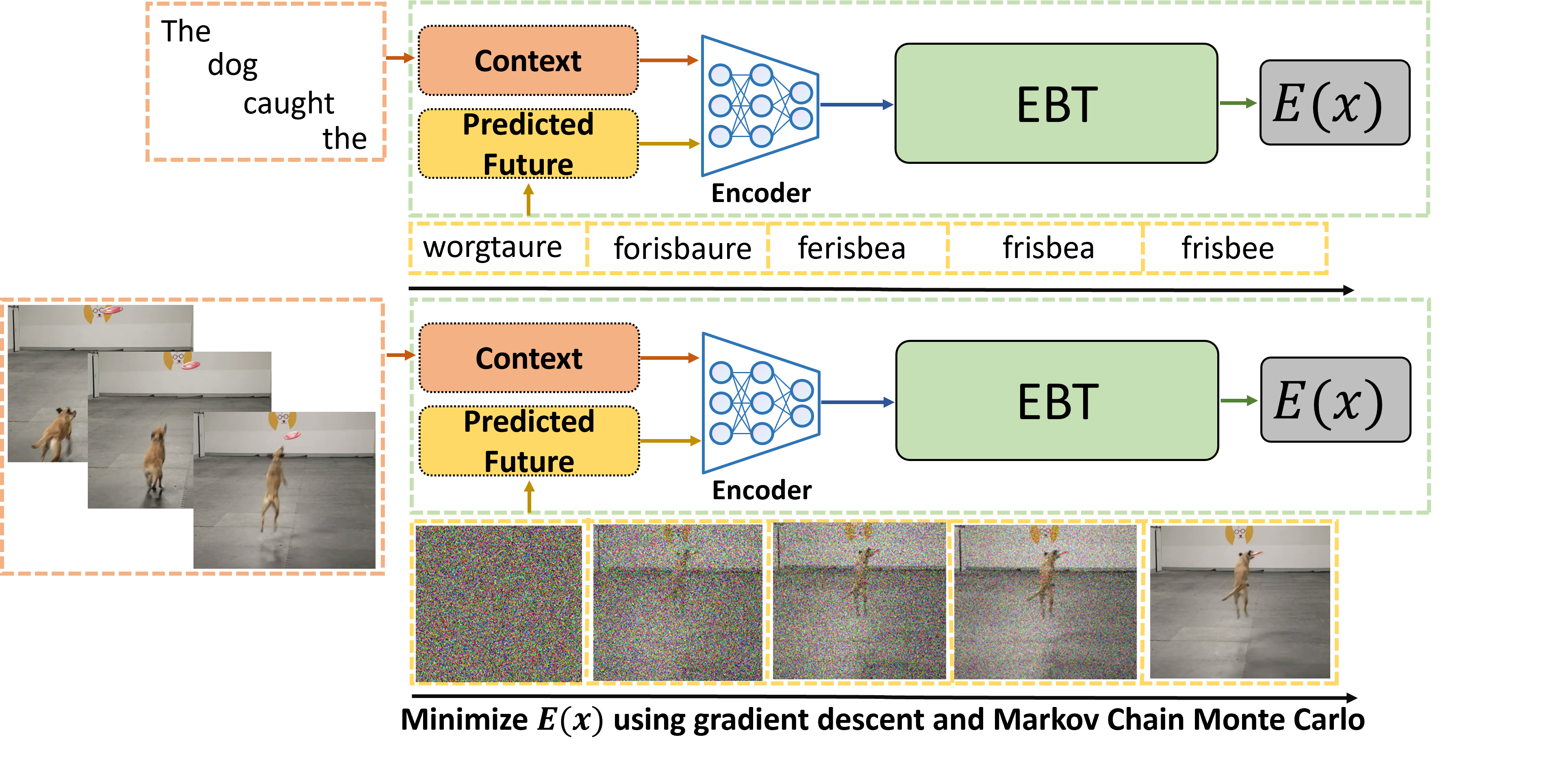}
    \caption{Architecture of {\pa} trained models for natural language processing and computer vision. MCMC is used to iteratively refine on the prediction of a future state until convergence of the predicted energy. Each yellow box corresponds to a different ``predicted future state'' based off of the current MCMC iteration (the first condition in this case is random).} 

    \label{fig:model_architecture}
\end{figure}

\subsection{Markov Chain Monte Carlo (MCMC) Approaches}
\label{sec:mcmc}
One useful property of {\pa} is the ability to leverage multiple forward passes in making a prediction. This is done through Markov Chain Monte Carlo (MCMC) methods, where an initial condition is passed into the model and the gradient of this condition with respect to the outputted energy is calculated (as the entire model is differentiable). Using this gradient, the input is updated. This is done iteratively until the convergence of the outputted energy (shown in Fig.~\ref{fig:model_architecture}). The usage of MCMC introduces a couple of new hyperparameters. First is the step size \(\alpha\), which is used to determine how much to multiply the gradient by when adjusting the predicted input. We make this a learnable parameter, but still find it important to tweak the initial value during training as it affects stability (Table \ref{tab:design_choices}). Two other hyperparameters introduced by MCMC are the learning rate multiplier of \(\alpha\) as well as the number of MCMC steps during training (this can be changed during inference). The values for these hyperparameters are detailed further in Section \ref{sec:training_details}. 

\begin{table*}[!t]
	\centering
 \caption{\textbf{Ablations on key {\pa} design choices in CV.} Convergent refers to whether the reconstruction loss converged.  Stable training means that there were no training loss spikes. The model achieved a similar loss with and without Langevin dynamics, so we elected to use the simpler alternative--without Langevin dynamics). Energy Loss and Bounds Loss are described in more detail in Section \ref{sec:supp_losses}. *In this case there were no loss spikes and the loss converged but only because the encoder experienced mode collapse.}
	\begin{tabular}{lcc}
  \toprule
		\textbf{Design Choice} & \hspace{-0.25cm}\textbf{Stable Training} & \hspace{-0.25cm}\textbf{Convergent} \\
  \toprule
		Energy Loss & \hspace{-0.25cm}\xmark & \hspace{-0.25cm}\cmark \\
		  Bounds Loss & \hspace{-0.25cm}\xmark & \hspace{-0.25cm}\cmark \\
            \midrule
		Unfrozen Encoder* & \hspace{-0.25cm}\xmark & \hspace{-0.25cm}\xmark \\
            Unclamped MCMC Gradient & \hspace{-0.25cm}\xmark & \hspace{-0.25cm}\xmark \\
		Non-Learnable $\alpha$ (MCMC Step Size) & \hspace{-0.25cm}\xmark & \hspace{-0.25cm}\cmark \\
            Lower Initial MCMC Step Size & \hspace{-0.25cm}\xmark & \hspace{-0.25cm}\cmark \\
            Langevin Dynamics & \hspace{-0.25cm}\cmark & \hspace{-0.25cm}\cmark \\
            Learnable Langevin Dynamics & \hspace{-0.25cm}\cmark & \hspace{-0.25cm}\cmark \\
  \midrule
		All Specified Design Choices & \hspace{-0.25cm}\cmark & \hspace{-0.25cm}\cmark \\
  \bottomrule
	\end{tabular}
 \label{tab:design_choices}
\end{table*}

\subsection{Energy-Based Transformer ({\parch})}
\label{sec:ebt_intro}

Two of the core mechanisms behind the success of the transformer are its attention mechanism and its parallelizability \cite{vaswani2017attention}. In the context of EBMs, the traditional transformer implementation poses a challenge due to the approximation of a joint distribution rather than a conditional distribution. To demonstrate why this poses a challenge, consider the case of the {\baseline} \(n \times n\) attention scores matrix after the causal mask has been applied:

\[
\text{scores} = 
\begin{bmatrix}
\alpha_{z_1,z_1} & 0 & \ldots & 0 \\
\alpha_{z_2,z_1} & \alpha_{z_2,z_2} & \ldots & 0 \\
\ldots & \ldots & \ldots & \ldots \\
\alpha_{z_n,z_1} & \alpha_{z_n,z_2} & \ldots & \alpha_{z_n,z_n} \\
\end{bmatrix},
\]

where \( \alpha_{z_i,z_j} \) represents the attention score (probability mass) from state \( z_i \) to state \( z_j \). Now, in the case of an EBM, where predictions of future states are made in the input space, the intended \(n \times n + 1 \) attention scores matrix would look like the following:

\begin{equation}
\text{scores} = 
\begin{bmatrix}
\alpha_{z_1,z_1} & \alpha_{z_1,\hat{z}_2} & 0 & \ldots & 0 \\
\alpha_{z_2,z_1} & \alpha_{z_2,z_2} & \alpha_{z_2,\hat{z}_3} & \ldots & 0 \\
\ldots & \ldots & \ldots & \ldots & \ldots \\
\alpha_{z_n,z_1} & \alpha_{z_n,z_2} & \alpha_{z_3,z_3} & \ldots & \alpha_{z_n,\hat{z}_{n+1}} \\
\end{bmatrix}.      \label{math:scores_ebm}
\end{equation}


This is challenging to compute because each \(\hat{z_i}\) along the superdiagonal is unique for its row. Consequently, this matrix cannot be computed with a matrix multiplication (\(\text{softmax}\left( \frac{QK^T}{\sqrt{d_k}} \right)\)) as in regular attention, as every value on the superdiagonal is a prediction and not a past state. 

Additionally, in a traditional transformer, if the context length is $n$, the size of the passed in tensor will be $bs \times n \times d$ where $bs$ is the batch size and $d$ is the embedding dimension. However, since EBMs learn a joint probability distribution, and thus make predictions in the input space, the input tensor needs to be different to allow for inputting future predictions. Therefore, for a context length of $n$, we define the first $n-1$ elements as \(z_o\), or the original sequence representations, and the final \(n-1\) elements as \(z_p\), or the predicted sequence representations.

The values for \(z_{o 1}^n\), or the given states, are computed the same as in the original transformer \cite{vaswani2017attention}, as the attention scores of the original states do not depend on the predicted states. This can be formalized as the following: \\
\begin{equation}
\text{Attention}(Q_o, K_o, V_o)_{z_{o 1}^n} = \text{softmax}\left(\frac{Q_oK_o^T}{\sqrt{d_k}}\right)V_0,
\end{equation}

Where $Q_o$, $K_o$, and $V_o$ are the Query, Key, and Value matrices of the past states \(z_{o 1}^n\). Every block of the transformer, the representations of all past states are updated in this manner, independent of the representations of the predicted future states. \\

For the computation of representations over future states, 3 matrices are also computed, but for the representations of predicted future states rather than past states. We call these $Q_p$, $K_p$, and $V_p$. First, we compute the self attention scores of all future representations to all past representations:

\begin{equation}
\text{unnormalized\_scores\_p} = \frac{Q_pK_o^T}{\sqrt{d_k}}. \label{eq:unnormalize_scores_p}
\end{equation}

Note, however, that the self-attention scores of each predicted future state with itself is not calculated--due to the key matrix being from the original states. Therefore, the superdiagonal needs to be replaced with the self-attention scores of each predicted future state with itself to achieve the attention score matrix shown in Equation \ref{math:scores_ebm}. The details of this implementation are provided in Section \ref{sec:ebt_full}.


\section{Experimentation}

\subsection{Computer Vision}

For all experiments below, we utilized a frozen DINOv2 backbone \cite{oquab2023dinov2} to encode all images into $768$ dimensional features. Then, using a reconstruction objective (the same one discussed in Section \ref{sec:energy}), the models are trained to predict the feature of the next image conditioned on all past features.

\begin{figure}
    \begin{center}
    \scriptsize
    \begin{tabular}{ccc}
        \hspace{-.3cm}\includegraphics[width=0.33\columnwidth]{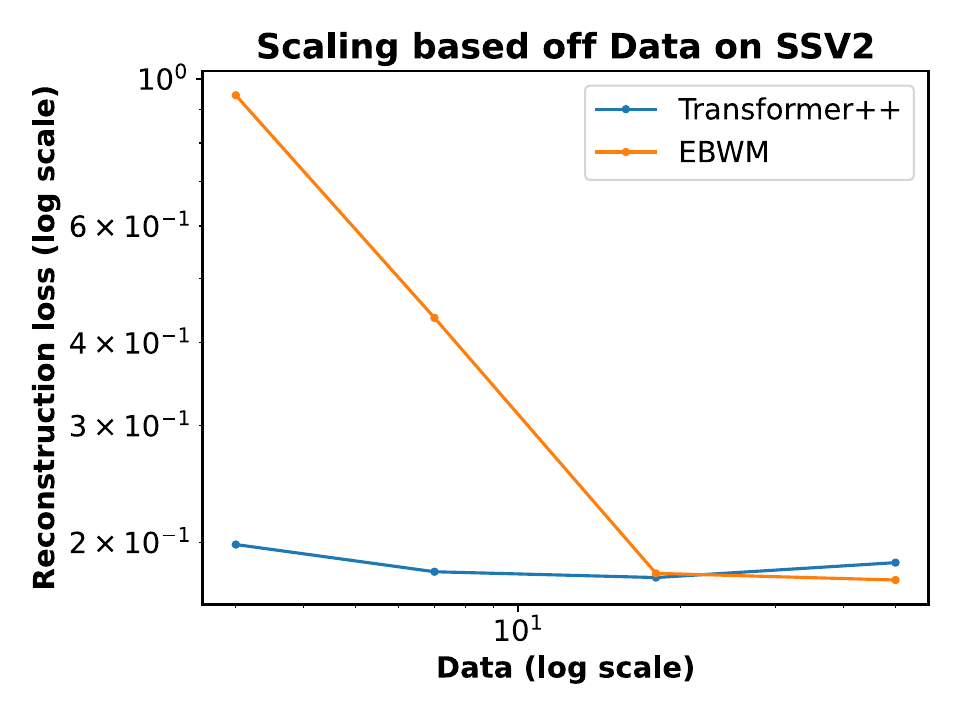} &
        \hspace{-.3cm}\includegraphics[width=0.33\columnwidth]{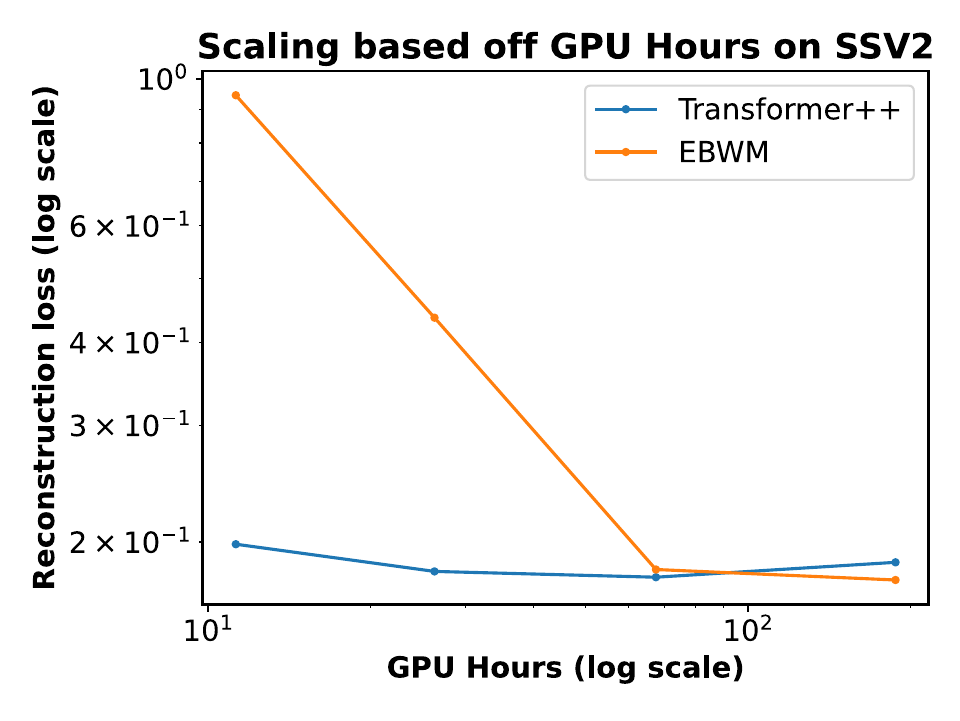} &
        \hspace{-.25cm}\includegraphics[width=0.33\columnwidth]{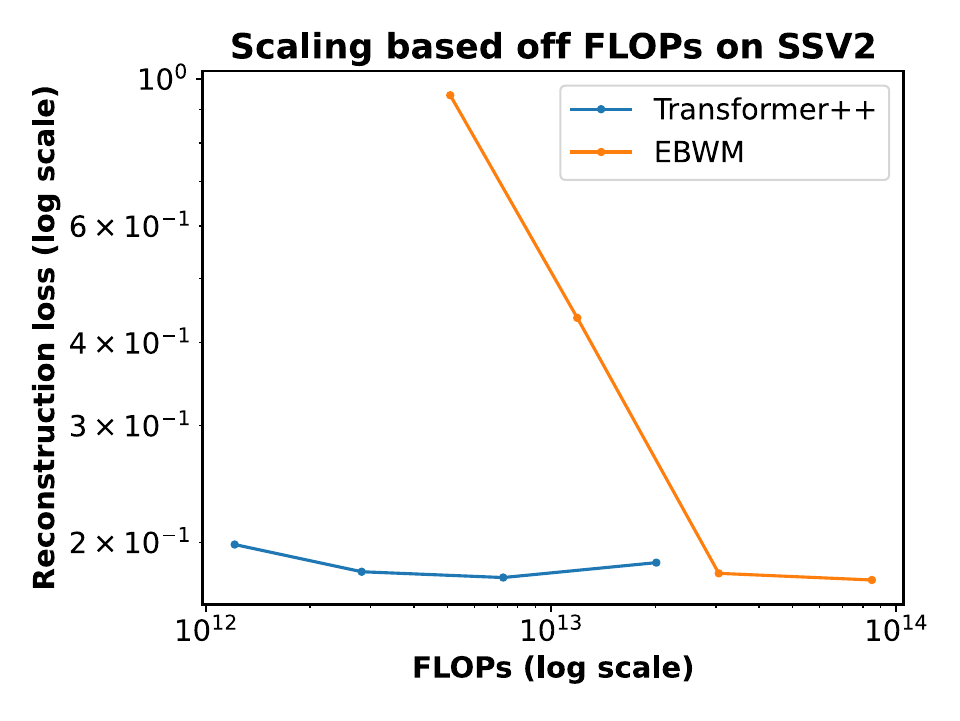} \\
        (a) Scaling for data across epochs. &
        (b) Scaling for A100 GPU Hours. &
        (c) Scaling for FLOPs. \\
    \end{tabular}

    \caption{The reconstruction loss across different scales of data, GPU hours, and FLOPs (lower and farther to the left curves are better). The third datapoint for {\baseline} was the minimum loss value achieved during training--meaning from then onwards the model overfits the data. Note that the rate the loss decreased for {\pa} is high initially due to being conditioned on random noise, whereas a simple copying of the most recent embedding for {\baseline} would yield a low reconstruction loss due to the nature of videos having high temporal consistency. Across runs, we consistently observe that {\pa} is less susceptible to overfitting. The lowest validation loss was achieved by {\pa}.}
    \label{fig:cv_scaling_ssv2}
    \end{center}
\end{figure}

\begin{figure}
    \begin{center}
    \scriptsize
    \begin{tabular}{ccc}
        \hspace{-.3cm}\includegraphics[width=0.318\columnwidth]{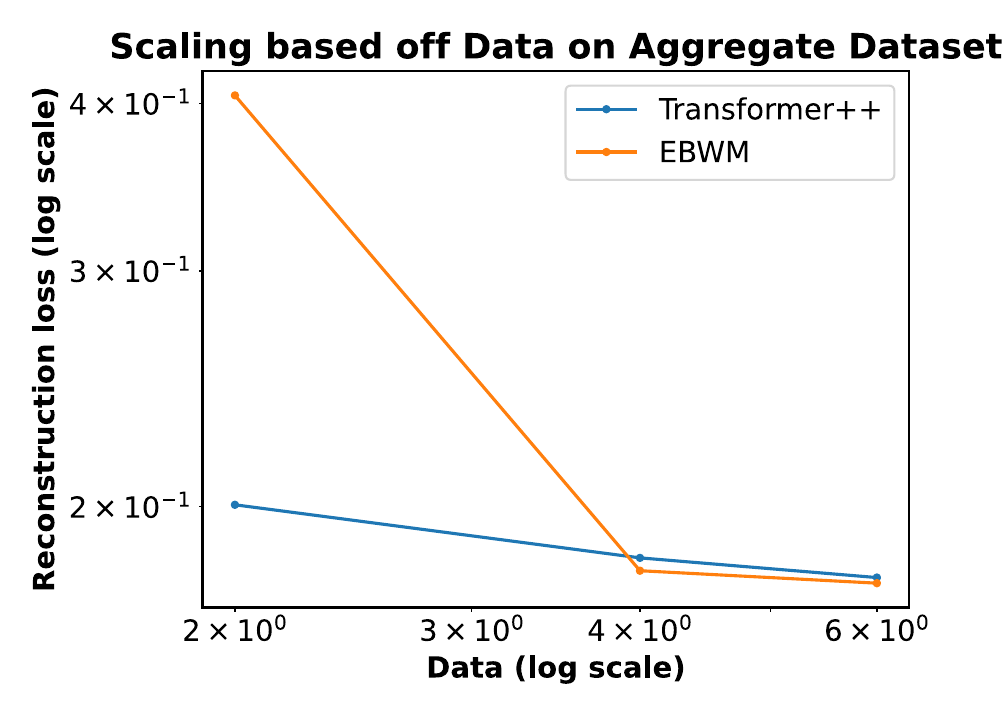} &
        \hspace{-.3cm}\includegraphics[width=0.34\columnwidth]{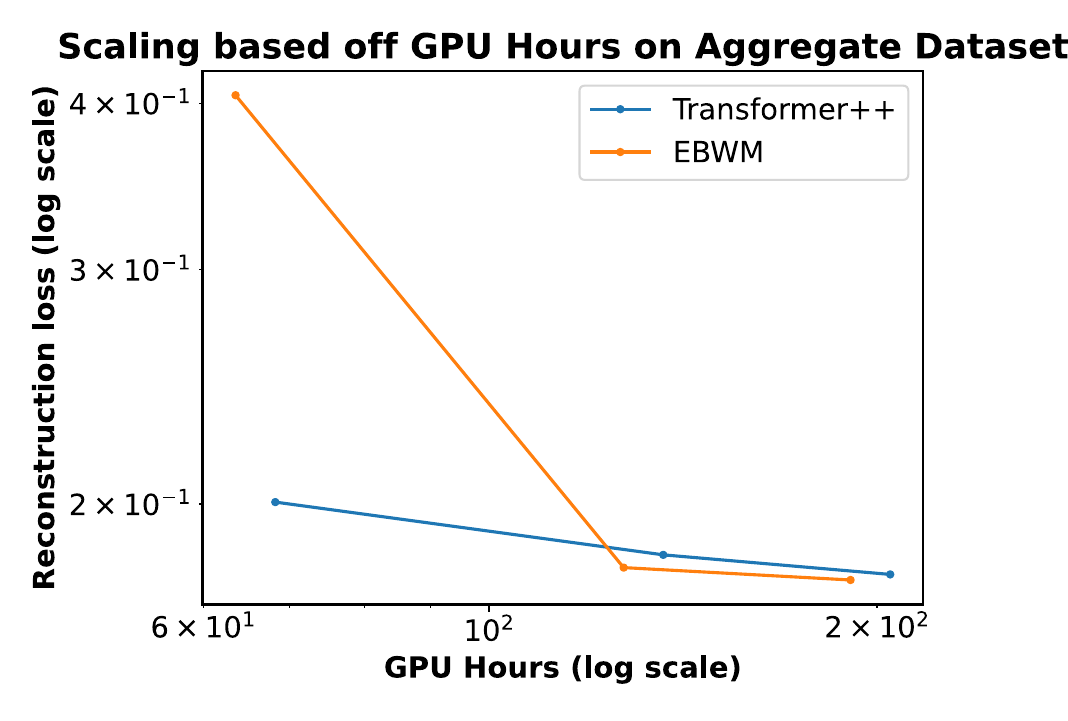} &
        \hspace{-.25cm}\includegraphics[width=0.326\columnwidth]{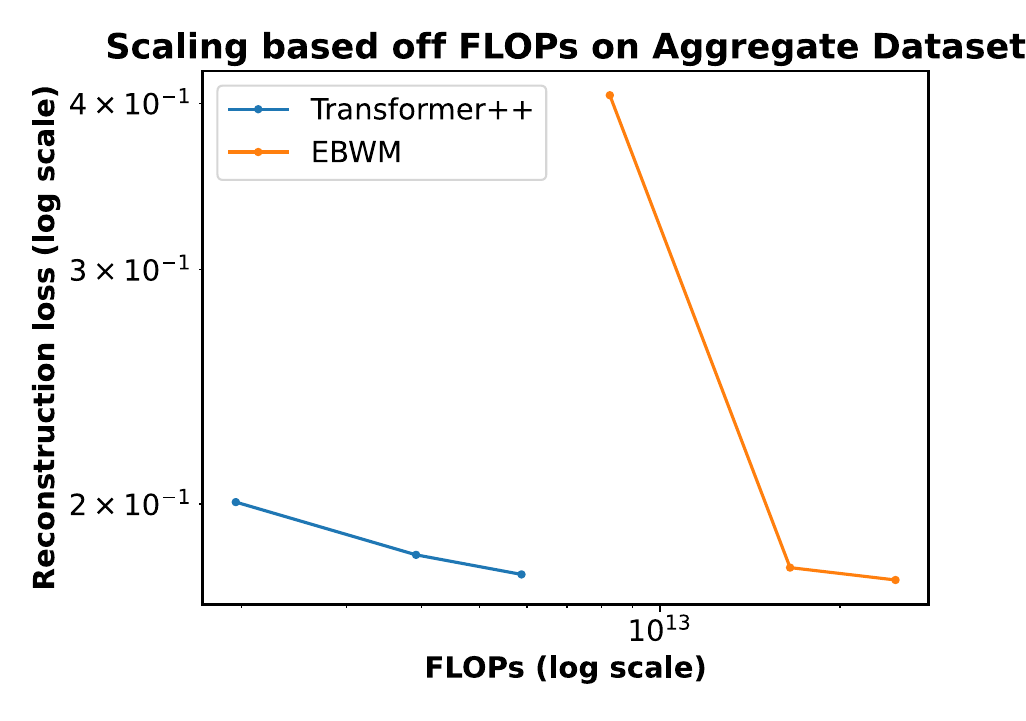} \\
        (a) Scaling for data across epochs. &
        (b) Scaling for A100 GPU Hours. &
        (c) Scaling for FLOPs. \\
    \end{tabular}
    
    \caption{The reconstruction loss across different scales of data, GPU hours, and FLOPs (lower and farther to the left curves are better) for the aggregated Kinetics-400 and SSV2 dataset. The results demonstrate the scalability of {\pa} when compared to {\baseline}, achieving better scaling in both data and compute hours.}
    \label{fig:cv_scaling_agg}
    \end{center}
\end{figure}

Scaling on the Something-Something V2 (SSV2) dataset \cite{goyal2017something} for data, GPU hours, and FLOPs are shown in Fig.~\ref{fig:cv_scaling_ssv2}. We observe {\baseline} overfitting past the third scale, therefore, we also investigate the scaling for data, GPU hours, and FLOPs on an aggregated dataset of Kinetics-400 \cite{kay2017kinetics} and SSV2 in Fig.~\ref{fig:cv_scaling_agg}. {\pa} consistently outperforms {\baseline} in data efficiency at higher scales, performs comparably or better in scaling with GPU hours, and performs worse in scaling with FLOPs. The experiments for NLP are in Section \ref{sec:nlp_exp}.

\section{Discussion}

Our experimental results demonstrate that initially {\pa} scales slower than {\baseline}, but as scale increases, it matches and eventually exceeds the performance of {\baseline} in data and GPU hour efficiency. This outcome is promising for higher compute regimes, as the scaling rate of {\pa} is higher than {\baseline} as computation increases. The slower initial scaling of {\pa} is primarily due to the requirement of learning to produce representations for generating gradients to predict the next state, which is more involved than directly predicting the next state. Additionally, as there is high temporal consistency in CV, simply copying the most recent embedding, which {\baseline} can easily do, yields decent performance. Throughout our experiments we see that {\pa} does not overfit the data, but {\baseline} do (Fig.~\ref{fig:cv_scaling_ssv2}). This reduced susceptibility to overfitting is due to EBMs learning a joint distribution, rather than a conditional distribution as in {\baseline}.

Although we do a side-by-side comparison of the scaling performance of {\pa} with {\baseline} throughout this work, we do not see {\pa} as a drop-in replacement for {\baseline}. Rather, having the four aspects of human cognition described, we see {\pa} as different and even complementary to {\baseline} (Section \ref{sec:complementary}). There exist several current real-world use cases, such as low-latency LLM serving, where doing a single forward pass is sufficient, and where the added inference overhead of backpropagating the gradient when using {\pa} would not be worth the extra computation. However, we also envision a world in which people needing long-term System 2 thinking to solve challenging problems use {\pa}. How much compute would it be worth dedicating to prove a long-standing mathematical conjecture? Similarly, in cases where a wide search is done, for example when solving challenging coding problems as in AlphaCode 2 \cite{Team_2023} and producing millions of potential solutions, using {\pa} for the evaluation and improvement of generated states could vastly improve performance. {\pa} offers a promising path towards neural architectures capable of achieving human-like cognition.

\section{Conclusion}

In this work, we proposed {\pa}, a novel training approach for autoregressive world models involving predictions made in the input space rather than in the output space through the usage of an EBM. Structuring the architecture in this manner offers distinct benefits for achieving human-like cognitive capabilities. We demonstrate the scalability of {\pa} through comprehensive performance comparisons with {\baseline} in both Vision and NLP. Based on our promising experimental results, we believe {\pa} provides a potential path towards achieving highly sought after human cognitive capabilities such as System 2 thinking or intelligent search. Our architecture has a few limitations that will be exciting future work to pursue:
\begin{itemize}
    \item \textbf{Additional hyperparameters.} Due to the usage of MCMC for generation, {\pa} involves additional hyperparameters such as the MCMC step size and the number of MCMC steps. 
    \item \textbf{Further Scaling.} Due to our limited computational resources, we did not investigate the scaling of {\pa} above \(1,000\) A100 GPU hour runs, leaving the scaling trends for larger training runs unexplored. However, the experiments conducted show the rate of scaling for {\pa} generally being higher than in {\baseline} as scale increases, offering promise towards further compute regimes.
\end{itemize}



{
    \small
    \bibliographystyle{unsrtnat}
    \bibliography{references}

\begin{thebibliography}{121}
\providecommand{\natexlab}[1]{#1}
\providecommand{\url}[1]{\texttt{#1}}
\expandafter\ifx\csname urlstyle\endcsname\relax
  \providecommand{\doi}[1]{doi: #1}\else
  \providecommand{\doi}{doi: \begingroup \urlstyle{rm}\Url}\fi

\bibitem[Gu and Dao(2023)]{gu2023mamba}
Albert Gu and Tri Dao.
\newblock Mamba: Linear-time sequence modeling with selective state spaces.
\newblock \emph{arXiv preprint arXiv:2312.00752}, 2023.

\bibitem[Peng et~al.(2023)Peng, Alcaide, Anthony, Albalak, Arcadinho, Cao, Cheng, Chung, Grella, GV, et~al.]{peng2023rwkv}
Bo~Peng, Eric Alcaide, Quentin Anthony, Alon Albalak, Samuel Arcadinho, Huanqi Cao, Xin Cheng, Michael Chung, Matteo Grella, Kranthi~Kiran GV, et~al.
\newblock Rwkv: Reinventing rnns for the transformer era.
\newblock \emph{arXiv preprint arXiv:2305.13048}, 2023.

\bibitem[Höppe et~al.(2022)Höppe, Mehrjou, Bauer, Nielsen, and Dittadi]{höppe2022diffusion}
Tobias Höppe, Arash Mehrjou, Stefan Bauer, Didrik Nielsen, and Andrea Dittadi.
\newblock Diffusion models for video prediction and infilling, 2022.

\bibitem[Sun et~al.(2023)Sun, Dong, Huang, Ma, Xia, Xue, Wang, and Wei]{sun2023retentive}
Yutao Sun, Li~Dong, Shaohan Huang, Shuming Ma, Yuqing Xia, Jilong Xue, Jianyong Wang, and Furu Wei.
\newblock Retentive network: A successor to transformer for large language models, 2023.

\bibitem[Hochreiter and Schmidhuber(1997)]{hochreiter1997long}
Sepp Hochreiter and J{\"u}rgen Schmidhuber.
\newblock Long short-term memory.
\newblock \emph{Neural computation}, 9\penalty0 (8):\penalty0 1735--1780, 1997.

\bibitem[Bardes et~al.(2023{\natexlab{a}})Bardes, Ponce, and LeCun]{bardes2023mcjepa}
Adrien Bardes, Jean Ponce, and Yann LeCun.
\newblock Mc-jepa: A joint-embedding predictive architecture for self-supervised learning of motion and content features, 2023{\natexlab{a}}.

\bibitem[Oquab et~al.(2023)Oquab, Darcet, Moutakanni, Vo, Szafraniec, Khalidov, Fernandez, Haziza, Massa, El-Nouby, Assran, Ballas, Galuba, Howes, Huang, Li, Misra, Rabbat, Sharma, Synnaeve, Xu, Jegou, Mairal, Labatut, Joulin, and Bojanowski]{oquab2023dinov2}
Maxime Oquab, Timothée Darcet, Théo Moutakanni, Huy Vo, Marc Szafraniec, Vasil Khalidov, Pierre Fernandez, Daniel Haziza, Francisco Massa, Alaaeldin El-Nouby, Mahmoud Assran, Nicolas Ballas, Wojciech Galuba, Russell Howes, Po-Yao Huang, Shang-Wen Li, Ishan Misra, Michael Rabbat, Vasu Sharma, Gabriel Synnaeve, Hu~Xu, Hervé Jegou, Julien Mairal, Patrick Labatut, Armand Joulin, and Piotr Bojanowski.
\newblock Dinov2: Learning robust visual features without supervision, 2023.

\bibitem[He et~al.(2021)He, Chen, Xie, Li, Dollár, and Girshick]{he2021masked}
Kaiming He, Xinlei Chen, Saining Xie, Yanghao Li, Piotr Dollár, and Ross Girshick.
\newblock Masked autoencoders are scalable vision learners, 2021.

\bibitem[Caron et~al.(2021)Caron, Touvron, Misra, Jégou, Mairal, Bojanowski, and Joulin]{caron2021emerging}
Mathilde Caron, Hugo Touvron, Ishan Misra, Hervé Jégou, Julien Mairal, Piotr Bojanowski, and Armand Joulin.
\newblock Emerging properties in self-supervised vision transformers, 2021.

\bibitem[He et~al.(2020)He, Fan, Wu, Xie, and Girshick]{he2020momentum}
Kaiming He, Haoqi Fan, Yuxin Wu, Saining Xie, and Ross Girshick.
\newblock Momentum contrast for unsupervised visual representation learning.
\newblock In \emph{Proceedings of the IEEE/CVF conference on computer vision and pattern recognition}, pages 9729--9738, 2020.

\bibitem[Chen et~al.(2020)Chen, Kornblith, Norouzi, and Hinton]{chen2020simple}
Ting Chen, Simon Kornblith, Mohammad Norouzi, and Geoffrey Hinton.
\newblock A simple framework for contrastive learning of visual representations, 2020.

\bibitem[OpenAI(2023{\natexlab{a}})]{openai2023gpt4}
OpenAI.
\newblock Gpt-4 technical report, 2023{\natexlab{a}}.

\bibitem[Zhao et~al.(2023)Zhao, Zhou, Li, Tang, Wang, Hou, Min, Zhang, Zhang, Dong, Du, Yang, Chen, Chen, Jiang, Ren, Li, Tang, Liu, Liu, Nie, and Wen]{zhao2023survey}
Wayne~Xin Zhao, Kun Zhou, Junyi Li, Tianyi Tang, Xiaolei Wang, Yupeng Hou, Yingqian Min, Beichen Zhang, Junjie Zhang, Zican Dong, Yifan Du, Chen Yang, Yushuo Chen, Zhipeng Chen, Jinhao Jiang, Ruiyang Ren, Yifan Li, Xinyu Tang, Zikang Liu, Peiyu Liu, Jian-Yun Nie, and Ji-Rong Wen.
\newblock A survey of large language models, 2023.

\bibitem[Brown et~al.(2020)Brown, Mann, Ryder, Subbiah, Kaplan, Dhariwal, Neelakantan, Shyam, Sastry, Askell, Agarwal, Herbert-Voss, Krueger, Henighan, Child, Ramesh, Ziegler, Wu, Winter, Hesse, Chen, Sigler, Litwin, Gray, Chess, Clark, Berner, McCandlish, Radford, Sutskever, and Amodei]{brown2020language}
Tom~B. Brown, Benjamin Mann, Nick Ryder, Melanie Subbiah, Jared Kaplan, Prafulla Dhariwal, Arvind Neelakantan, Pranav Shyam, Girish Sastry, Amanda Askell, Sandhini Agarwal, Ariel Herbert-Voss, Gretchen Krueger, Tom Henighan, Rewon Child, Aditya Ramesh, Daniel~M. Ziegler, Jeffrey Wu, Clemens Winter, Christopher Hesse, Mark Chen, Eric Sigler, Mateusz Litwin, Scott Gray, Benjamin Chess, Jack Clark, Christopher Berner, Sam McCandlish, Alec Radford, Ilya Sutskever, and Dario Amodei.
\newblock Language models are few-shot learners, 2020.

\bibitem[Alayrac et~al.(2022)Alayrac, Donahue, Luc, Miech, Barr, Hasson, Lenc, Mensch, Millican, Reynolds, Ring, Rutherford, Cabi, Han, Gong, Samangooei, Monteiro, Menick, Borgeaud, Brock, Nematzadeh, Sharifzadeh, Binkowski, Barreira, Vinyals, Zisserman, and Simonyan]{alayrac2022flamingo}
Jean-Baptiste Alayrac, Jeff Donahue, Pauline Luc, Antoine Miech, Iain Barr, Yana Hasson, Karel Lenc, Arthur Mensch, Katie Millican, Malcolm Reynolds, Roman Ring, Eliza Rutherford, Serkan Cabi, Tengda Han, Zhitao Gong, Sina Samangooei, Marianne Monteiro, Jacob Menick, Sebastian Borgeaud, Andrew Brock, Aida Nematzadeh, Sahand Sharifzadeh, Mikolaj Binkowski, Ricardo Barreira, Oriol Vinyals, Andrew Zisserman, and Karen Simonyan.
\newblock Flamingo: a visual language model for few-shot learning, 2022.

\bibitem[Devlin et~al.(2019)Devlin, Chang, Lee, and Toutanova]{devlin2019bert}
Jacob Devlin, Ming-Wei Chang, Kenton Lee, and Kristina Toutanova.
\newblock Bert: Pre-training of deep bidirectional transformers for language understanding, 2019.

\bibitem[Touvron et~al.(2023)Touvron, Martin, Stone, Albert, Almahairi, Babaei, Bashlykov, Batra, Bhargava, Bhosale, Bikel, Blecher, Ferrer, Chen, Cucurull, Esiobu, Fernandes, Fu, Fu, Fuller, Gao, Goswami, Goyal, Hartshorn, Hosseini, Hou, Inan, Kardas, Kerkez, Khabsa, Kloumann, Korenev, Koura, Lachaux, Lavril, Lee, Liskovich, Lu, Mao, Martinet, Mihaylov, Mishra, Molybog, Nie, Poulton, Reizenstein, Rungta, Saladi, Schelten, Silva, Smith, Subramanian, Tan, Tang, Taylor, Williams, Kuan, Xu, Yan, Zarov, Zhang, Fan, Kambadur, Narang, Rodriguez, Stojnic, Edunov, and Scialom]{touvron2023llama}
Hugo Touvron, Louis Martin, Kevin Stone, Peter Albert, Amjad Almahairi, Yasmine Babaei, Nikolay Bashlykov, Soumya Batra, Prajjwal Bhargava, Shruti Bhosale, Dan Bikel, Lukas Blecher, Cristian~Canton Ferrer, Moya Chen, Guillem Cucurull, David Esiobu, Jude Fernandes, Jeremy Fu, Wenyin Fu, Brian Fuller, Cynthia Gao, Vedanuj Goswami, Naman Goyal, Anthony Hartshorn, Saghar Hosseini, Rui Hou, Hakan Inan, Marcin Kardas, Viktor Kerkez, Madian Khabsa, Isabel Kloumann, Artem Korenev, Punit~Singh Koura, Marie-Anne Lachaux, Thibaut Lavril, Jenya Lee, Diana Liskovich, Yinghai Lu, Yuning Mao, Xavier Martinet, Todor Mihaylov, Pushkar Mishra, Igor Molybog, Yixin Nie, Andrew Poulton, Jeremy Reizenstein, Rashi Rungta, Kalyan Saladi, Alan Schelten, Ruan Silva, Eric~Michael Smith, Ranjan Subramanian, Xiaoqing~Ellen Tan, Binh Tang, Ross Taylor, Adina Williams, Jian~Xiang Kuan, Puxin Xu, Zheng Yan, Iliyan Zarov, Yuchen Zhang, Angela Fan, Melanie Kambadur, Sharan Narang, Aurelien Rodriguez, Robert Stojnic, Sergey Edunov, and Thomas
  Scialom.
\newblock Llama 2: Open foundation and fine-tuned chat models, 2023.

\bibitem[Benita et~al.(2023)Benita, Elad, and Keshet]{benita2023diffar}
Roi Benita, Michael Elad, and Joseph Keshet.
\newblock Diffar: Denoising diffusion autoregressive model for raw speech waveform generation.
\newblock \emph{arXiv preprint arXiv:2310.01381}, 2023.

\bibitem[Hsu et~al.(2023)Hsu, Liu, Liu, and Lee]{hsu2023parallel}
Po-chun Hsu, Da-Rong Liu, Andy~T Liu, and Hung-yi Lee.
\newblock Parallel synthesis for autoregressive speech generation.
\newblock \emph{IEEE/ACM Transactions on Audio, Speech, and Language Processing}, 2023.

\bibitem[Borsos et~al.(2023)Borsos, Marinier, Vincent, Kharitonov, Pietquin, Sharifi, Roblek, Teboul, Grangier, Tagliasacchi, and Zeghidour]{borsos2023audiolm}
Zalán Borsos, Raphaël Marinier, Damien Vincent, Eugene Kharitonov, Olivier Pietquin, Matt Sharifi, Dominik Roblek, Olivier Teboul, David Grangier, Marco Tagliasacchi, and Neil Zeghidour.
\newblock Audiolm: a language modeling approach to audio generation, 2023.

\bibitem[Noy and Zhang(2023)]{noy2023experimental}
Shakked Noy and Whitney Zhang.
\newblock Experimental evidence on the productivity effects of generative artificial intelligence.
\newblock \emph{Science}, 381\penalty0 (6654):\penalty0 187--192, 2023.

\bibitem[Spataro(2023)]{Spataro_2023}
Jared Spataro.
\newblock Introducing microsoft 365 copilot – your copilot for work, March 2023.
\newblock URL \url{https://blogs.microsoft.com/blog/2023/03/16/introducing-microsoft-365-copilot-your-copilot-for-work/}.

\bibitem[sta()]{stablevideodiffusion}
URL \url{https://stability.ai/news/stable-video-diffusion-open-ai-video-model}.

\bibitem[Hey()]{HeySiri}
URL \url{https://machinelearning.apple.com/research/hey-siri}.

\bibitem[OpenAI(2023{\natexlab{b}})]{OpenAI_2023}
OpenAI.
\newblock Chatgpt can now see, hear, and speak, September 2023{\natexlab{b}}.
\newblock URL \url{https://openai.com/index/chatgpt-can-now-see-hear-and-speak/}.

\bibitem[Radford et~al.(2019)Radford, Wu, Child, Luan, Amodei, Sutskever, et~al.]{radford2019language}
Alec Radford, Jeffrey Wu, Rewon Child, David Luan, Dario Amodei, Ilya Sutskever, et~al.
\newblock Language models are unsupervised multitask learners.
\newblock \emph{OpenAI blog}, 1\penalty0 (8):\penalty0 9, 2019.

\bibitem[Mahjourian et~al.(2017)Mahjourian, Wicke, and Angelova]{mahjourian2017geometry}
Reza Mahjourian, Martin Wicke, and Anelia Angelova.
\newblock Geometry-based next frame prediction from monocular video.
\newblock In \emph{2017 IEEE Intelligent Vehicles Symposium (IV)}, pages 1700--1707. IEEE, 2017.

\bibitem[Zhou et~al.(2020)Zhou, Dong, and El~Saddik]{zhou2020deep}
Yufan Zhou, Haiwei Dong, and Abdulmotaleb El~Saddik.
\newblock Deep learning in next-frame prediction: A benchmark review.
\newblock \emph{IEEE Access}, 8:\penalty0 69273--69283, 2020.

\bibitem[Oord et~al.(2018)Oord, Li, and Vinyals]{oord2018representation}
Aaron van~den Oord, Yazhe Li, and Oriol Vinyals.
\newblock Representation learning with contrastive predictive coding.
\newblock \emph{arXiv preprint arXiv:1807.03748}, 2018.

\bibitem[Islam et~al.(2023{\natexlab{a}})Islam, Gladstone, Islam, and Iqbal]{islam2023eqa}
Md~Mofijul Islam, Alexi Gladstone, Riashat Islam, and Tariq Iqbal.
\newblock Eqa-mx: Embodied question answering using multimodal expression.
\newblock In \emph{The Twelfth International Conference on Learning Representations}, 2023{\natexlab{a}}.

\bibitem[Van Den~Oord et~al.(2017)Van Den~Oord, Vinyals, et~al.]{van2017neural}
Aaron Van Den~Oord, Oriol Vinyals, et~al.
\newblock Neural discrete representation learning.
\newblock \emph{Advances in neural information processing systems}, 30, 2017.

\bibitem[Bai et~al.(2023)Bai, Geng, Mangalam, Bar, Yuille, Darrell, Malik, and Efros]{bai2023sequential}
Yutong Bai, Xinyang Geng, Karttikeya Mangalam, Amir Bar, Alan Yuille, Trevor Darrell, Jitendra Malik, and Alexei~A Efros.
\newblock Sequential modeling enables scalable learning for large vision models.
\newblock \emph{arXiv preprint arXiv:2312.00785}, 2023.

\bibitem[Hu et~al.(2023)Hu, Russell, Yeo, Murez, Fedoseev, Kendall, Shotton, and Corrado]{hu2023gaia1}
Anthony Hu, Lloyd Russell, Hudson Yeo, Zak Murez, George Fedoseev, Alex Kendall, Jamie Shotton, and Gianluca Corrado.
\newblock Gaia-1: A generative world model for autonomous driving, 2023.

\bibitem[Islam et~al.(2022{\natexlab{a}})Islam, Zang, Tomar, Didolkar, Islam, Arnob, Iqbal, Li, Goyal, Heess, et~al.]{islam2022representation}
Riashat Islam, Hongyu Zang, Manan Tomar, Aniket Didolkar, Md~Mofijul Islam, Samin~Yeasar Arnob, Tariq Iqbal, Xin Li, Anirudh Goyal, Nicolas Heess, et~al.
\newblock Representation learning in deep rl via discrete information bottleneck.
\newblock \emph{arXiv preprint arXiv:2212.13835}, 2022{\natexlab{a}}.

\bibitem[Yasar and Iqbal(2023)]{yasar2023vader}
Mohammad~Samin Yasar and Tariq Iqbal.
\newblock Vader: Vector-quantized generative adversarial network for motion prediction.
\newblock In \emph{2023 IEEE/RSJ International Conference on Intelligent Robots and Systems (IROS)}, pages 3827--3834. IEEE, 2023.

\bibitem[Lee et~al.(2021)Lee, Chen, Wang, Gu, Popuri, Ma, Polyak, Adi, He, Tang, et~al.]{lee2021direct}
Ann Lee, Peng-Jen Chen, Changhan Wang, Jiatao Gu, Sravya Popuri, Xutai Ma, Adam Polyak, Yossi Adi, Qing He, Yun Tang, et~al.
\newblock Direct speech-to-speech translation with discrete units.
\newblock \emph{arXiv preprint arXiv:2107.05604}, 2021.

\bibitem[Valle et~al.(2020)Valle, Shih, Prenger, and Catanzaro]{valle2020flowtron}
Rafael Valle, Kevin Shih, Ryan Prenger, and Bryan Catanzaro.
\newblock Flowtron: an autoregressive flow-based generative network for text-to-speech synthesis.
\newblock \emph{arXiv preprint arXiv:2005.05957}, 2020.

\bibitem[Hawthorne et~al.(2022)Hawthorne, Jaegle, Cangea, Borgeaud, Nash, Malinowski, Dieleman, Vinyals, Botvinick, Simon, Sheahan, Zeghidour, Alayrac, Carreira, and Engel]{hawthorne2022generalpurpose}
Curtis Hawthorne, Andrew Jaegle, Cătălina Cangea, Sebastian Borgeaud, Charlie Nash, Mateusz Malinowski, Sander Dieleman, Oriol Vinyals, Matthew Botvinick, Ian Simon, Hannah Sheahan, Neil Zeghidour, Jean-Baptiste Alayrac, João Carreira, and Jesse Engel.
\newblock General-purpose, long-context autoregressive modeling with perceiver ar, 2022.

\bibitem[OpenAI(2024)]{OpenAI_sora}
OpenAI.
\newblock Sora: first impressions, March 2024.
\newblock URL \url{https://openai.com/index/sora-first-impressions/}.

\bibitem[Valmeekam et~al.(2022)Valmeekam, Olmo, Sreedharan, and Kambhampati]{valmeekam2022large}
Karthik Valmeekam, Alberto Olmo, Sarath Sreedharan, and Subbarao Kambhampati.
\newblock Large language models still can't plan (a benchmark for llms on planning and reasoning about change).
\newblock \emph{arXiv preprint arXiv:2206.10498}, 2022.

\bibitem[Zhao and Zhang(2024)]{zhao2024exploring}
Jinman Zhao and Xueyan Zhang.
\newblock Exploring the limitations of large language models in compositional relation reasoning.
\newblock \emph{arXiv preprint arXiv:2403.02615}, 2024.

\bibitem[Li et~al.(2024)Li, Huang, Lu, Xiao, Pei, and Bai]{li2024survey}
Chengxuan Li, Di~Huang, Zeyu Lu, Yang Xiao, Qingqi Pei, and Lei Bai.
\newblock A survey on long video generation: Challenges, methods, and prospects.
\newblock \emph{arXiv preprint arXiv:2403.16407}, 2024.

\bibitem[Liu et~al.(2023)Liu, Jiang, Zhang, Liu, Zhang, Biswas, and Stone]{liu2023llm}
Bo~Liu, Yuqian Jiang, Xiaohan Zhang, Qiang Liu, Shiqi Zhang, Joydeep Biswas, and Peter Stone.
\newblock Llm+ p: Empowering large language models with optimal planning proficiency.
\newblock \emph{arXiv preprint arXiv:2304.11477}, 2023.

\bibitem[Mondorf and Plank(2024)]{mondorf2024beyond}
Philipp Mondorf and Barbara Plank.
\newblock Beyond accuracy: Evaluating the reasoning behavior of large language models--a survey.
\newblock \emph{arXiv preprint arXiv:2404.01869}, 2024.

\bibitem[Agrawal(2010)]{agrawal2010study}
Kush Agrawal.
\newblock To study the phenomenon of the moravec's paradox.
\newblock \emph{arXiv preprint arXiv:1012.3148}, 2010.

\bibitem[West et~al.(2023)West, Lu, Dziri, Brahman, Li, Hwang, Jiang, Fisher, Ravichander, Chandu, et~al.]{west2023generative}
Peter West, Ximing Lu, Nouha Dziri, Faeze Brahman, Linjie Li, Jena~D Hwang, Liwei Jiang, Jillian Fisher, Abhilasha Ravichander, Khyathi Chandu, et~al.
\newblock The generative ai paradox:" what it can create, it may not understand".
\newblock \emph{arXiv preprint arXiv:2311.00059}, 2023.

\bibitem[Alexander(2003)]{alexander2003development}
Patricia~A Alexander.
\newblock The development of expertise: The journey from acclimation to proficiency.
\newblock \emph{Educational researcher}, 32\penalty0 (8):\penalty0 10--14, 2003.

\bibitem[Bubic et~al.(2010)Bubic, Von~Cramon, and Schubotz]{bubic2010prediction}
Andreja Bubic, D~Yves Von~Cramon, and Ricarda~I Schubotz.
\newblock Prediction, cognition and the brain.
\newblock \emph{Frontiers in human neuroscience}, 4:\penalty0 1094, 2010.

\bibitem[Tomasi et~al.(2015)Tomasi, Wang, Studentsova, and Volkow]{Tomasi2015Dissecting}
D.~Tomasi, Gene-Jack Wang, Y.~Studentsova, and N.~Volkow.
\newblock Dissecting neural responses to temporal prediction, attention, and memory: Effects of reward learning and interoception on time perception.
\newblock \emph{Cerebral cortex}, 25 10:\penalty0 3856--67, 2015.
\newblock \doi{10.1093/cercor/bhu269}.

\bibitem[Schacter et~al.(2007)Schacter, Addis, and Buckner]{Schacter2007Remembering}
D.~Schacter, D.~Addis, and R.~Buckner.
\newblock Remembering the past to imagine the future: the prospective brain.
\newblock \emph{Nature Reviews Neuroscience}, 8:\penalty0 657--661, 2007.
\newblock \doi{10.1038/nrn2213}.

\bibitem[Huang and Rao(2011)]{huang2011predictive}
Yanping Huang and Rajesh~PN Rao.
\newblock Predictive coding.
\newblock \emph{Wiley Interdisciplinary Reviews: Cognitive Science}, 2\penalty0 (5):\penalty0 580--593, 2011.

\bibitem[Nayebi et~al.(2023)Nayebi, Rajalingham, Jazayeri, and Yang]{Nayebi2023Neural}
Aran Nayebi, R.~Rajalingham, M.~Jazayeri, and G.~R. Yang.
\newblock Neural foundations of mental simulation: Future prediction of latent representations on dynamic scenes.
\newblock \emph{ArXiv}, 2023.
\newblock \doi{10.48550/arXiv.2305.11772}.

\bibitem[Connell and Keane(2006)]{Connell2006A}
L.~Connell and Mark~T. Keane.
\newblock A model of plausibility.
\newblock \emph{Cognitive science}, 30 1:\penalty0 95--120, 2006.
\newblock \doi{10.1207/s15516709cog0000_53}.

\bibitem[Brown and Brüne(2012)]{Brown2012The}
E.~Brown and M.~Brüne.
\newblock The role of prediction in social neuroscience.
\newblock \emph{Frontiers in Human Neuroscience}, 6, 2012.
\newblock \doi{10.3389/fnhum.2012.00147}.

\bibitem[Kahneman(2011)]{kahneman2011thinking}
Daniel Kahneman.
\newblock \emph{Thinking, fast and slow}.
\newblock macmillan, 2011.

\bibitem[Fuster(1991)]{fuster1991prefrontal}
Joaquin~M Fuster.
\newblock The prefrontal cortex and its relation to behavior.
\newblock \emph{Progress in brain research}, 87:\penalty0 201--211, 1991.

\bibitem[Rougier et~al.(2005)Rougier, Noelle, Braver, Cohen, and O'Reilly]{rougier2005prefrontal}
Nicolas~P Rougier, David~C Noelle, Todd~S Braver, Jonathan~D Cohen, and Randall~C O'Reilly.
\newblock Prefrontal cortex and flexible cognitive control: Rules without symbols.
\newblock \emph{Proceedings of the National Academy of Sciences}, 102\penalty0 (20):\penalty0 7338--7343, 2005.

\bibitem[Tomani et~al.(2024)Tomani, Chaudhuri, Evtimov, Cremers, and Ibrahim]{tomani2024uncertainty}
Christian Tomani, Kamalika Chaudhuri, Ivan Evtimov, Daniel Cremers, and Mark Ibrahim.
\newblock Uncertainty-based abstention in llms improves safety and reduces hallucinations.
\newblock \emph{arXiv preprint arXiv:2404.10960}, 2024.

\bibitem[Peters et~al.(2017)Peters, McEwen, and Friston]{Peters2017Uncertainty}
A.~Peters, B.~McEwen, and Karl~J. Friston.
\newblock Uncertainty and stress: Why it causes diseases and how it is mastered by the brain.
\newblock \emph{Progress in Neurobiology}, 156:\penalty0 164--188, 2017.
\newblock \doi{10.1016/j.pneurobio.2017.05.004}.

\bibitem[Vilares et~al.(2012)Vilares, Howard, Fernandes, Gottfried, and Kording]{Vilares2012Differential}
I.~Vilares, J.~D. Howard, Hugo~L. Fernandes, J.~Gottfried, and Konrad~Paul Kording.
\newblock Differential representations of prior and likelihood uncertainty in the human brain.
\newblock \emph{Current Biology}, 22:\penalty0 1641--1648, 2012.
\newblock \doi{10.1016/j.cub.2012.07.010}.

\bibitem[Sarinopoulos et~al.(2010)Sarinopoulos, Grupe, Mackiewicz, Herrington, Lor, Steege, and Nitschke]{Sarinopoulos2010Uncertainty}
Issidoros~C. Sarinopoulos, D.~Grupe, Kristen~L. Mackiewicz, J.~Herrington, M.~Lor, E.~E. Steege, and J.~Nitschke.
\newblock Uncertainty during anticipation modulates neural responses to aversion in human insula and amygdala.
\newblock \emph{Cerebral cortex}, 20 4:\penalty0 929--40, 2010.
\newblock \doi{10.1093/cercor/bhp155}.

\bibitem[Vaswani et~al.(2017)Vaswani, Shazeer, Parmar, Uszkoreit, Jones, Gomez, Kaiser, and Polosukhin]{vaswani2017attention}
Ashish Vaswani, Noam Shazeer, Niki Parmar, Jakob Uszkoreit, Llion Jones, Aidan~N Gomez, {\L}ukasz Kaiser, and Illia Polosukhin.
\newblock Attention is all you need.
\newblock \emph{Advances in neural information processing systems}, 30, 2017.

\bibitem[LeCun(2022)]{lecun2022path}
Yann LeCun.
\newblock A path towards autonomous machine intelligence version 0.9. 2, 2022-06-27.
\newblock \emph{Open Review}, 62, 2022.

\bibitem[LeCun(2024)]{lecunworldmodeldef}
Yann LeCun, 2024.
\newblock URL \url{https://www.linkedin.com/posts/yann-lecun_lots-of-confusion-about-what-a-world-model-activity-7165738293223931904-vdgR}.

\bibitem[Ha and Schmidhuber(2018)]{ha2018world}
David Ha and J{\"u}rgen Schmidhuber.
\newblock World models.
\newblock \emph{arXiv preprint arXiv:1803.10122}, 2018.

\bibitem[Liu et~al.(2024{\natexlab{a}})Liu, Yan, Zaharia, and Abbeel]{liu2024world}
Hao Liu, Wilson Yan, Matei Zaharia, and Pieter Abbeel.
\newblock World model on million-length video and language with ringattention.
\newblock \emph{arXiv preprint arXiv:2402.08268}, 2024{\natexlab{a}}.

\bibitem[Morari and Lee(1999)]{morari1999model}
Manfred Morari and Jay~H Lee.
\newblock Model predictive control: past, present and future.
\newblock \emph{Computers \& chemical engineering}, 23\penalty0 (4-5):\penalty0 667--682, 1999.

\bibitem[Taniguchi et~al.(2023)Taniguchi, Murata, Suzuki, Ognibene, Lanillos, Ugur, Jamone, Nakamura, Ciria, Lara, et~al.]{taniguchi2023world}
Tadahiro Taniguchi, Shingo Murata, Masahiro Suzuki, Dimitri Ognibene, Pablo Lanillos, Emre Ugur, Lorenzo Jamone, Tomoaki Nakamura, Alejandra Ciria, Bruno Lara, et~al.
\newblock World models and predictive coding for cognitive and developmental robotics: frontiers and challenges.
\newblock \emph{Advanced Robotics}, 37\penalty0 (13):\penalty0 780--806, 2023.

\bibitem[Jiang et~al.(2023)Jiang, Sablayrolles, Mensch, Bamford, Chaplot, Casas, Bressand, Lengyel, Lample, Saulnier, et~al.]{jiang2023mistral}
Albert~Q Jiang, Alexandre Sablayrolles, Arthur Mensch, Chris Bamford, Devendra~Singh Chaplot, Diego de~las Casas, Florian Bressand, Gianna Lengyel, Guillaume Lample, Lucile Saulnier, et~al.
\newblock Mistral 7b.
\newblock \emph{arXiv preprint arXiv:2310.06825}, 2023.

\bibitem[Team et~al.(2023)Team, Anil, Borgeaud, Wu, Alayrac, Yu, Soricut, Schalkwyk, Dai, Hauth, et~al.]{team2023gemini}
Gemini Team, Rohan Anil, Sebastian Borgeaud, Yonghui Wu, Jean-Baptiste Alayrac, Jiahui Yu, Radu Soricut, Johan Schalkwyk, Andrew~M Dai, Anja Hauth, et~al.
\newblock Gemini: a family of highly capable multimodal models.
\newblock \emph{arXiv preprint arXiv:2312.11805}, 2023.

\bibitem[Bardes et~al.(2023{\natexlab{b}})Bardes, Garrido, Ponce, Chen, Rabbat, LeCun, Assran, and Ballas]{bardes2023v}
Adrien Bardes, Quentin Garrido, Jean Ponce, Xinlei Chen, Michael Rabbat, Yann LeCun, Mido Assran, and Nicolas Ballas.
\newblock V-jepa: Latent video prediction for visual representation learning.
\newblock 2023{\natexlab{b}}.

\bibitem[H{\"o}ppe et~al.(2022)H{\"o}ppe, Mehrjou, Bauer, Nielsen, and Dittadi]{hoppe2022diffusion}
Tobias H{\"o}ppe, Arash Mehrjou, Stefan Bauer, Didrik Nielsen, and Andrea Dittadi.
\newblock Diffusion models for video prediction and infilling.
\newblock \emph{arXiv preprint arXiv:2206.07696}, 2022.

\bibitem[Weissenborn et~al.(2019)Weissenborn, T{\"a}ckstr{\"o}m, and Uszkoreit]{weissenborn2019scaling}
Dirk Weissenborn, Oscar T{\"a}ckstr{\"o}m, and Jakob Uszkoreit.
\newblock Scaling autoregressive video models.
\newblock \emph{arXiv preprint arXiv:1906.02634}, 2019.

\bibitem[Lu et~al.(2023)Lu, Clark, Lee, Zhang, Khosla, Marten, Hoiem, and Kembhavi]{lu2023unifiedio}
Jiasen Lu, Christopher Clark, Sangho Lee, Zichen Zhang, Savya Khosla, Ryan Marten, Derek Hoiem, and Aniruddha Kembhavi.
\newblock Unified-io 2: Scaling autoregressive multimodal models with vision, language, audio, and action, 2023.

\bibitem[Chen et~al.(2022)Chen, Wu, Yoon, and Ahn]{chen2022transdreamer}
Chang Chen, Yi-Fu Wu, Jaesik Yoon, and Sungjin Ahn.
\newblock Transdreamer: Reinforcement learning with transformer world models.
\newblock \emph{arXiv preprint arXiv:2202.09481}, 2022.

\bibitem[Hafner et~al.(2019)Hafner, Lillicrap, Ba, and Norouzi]{hafner2019dream}
Danijar Hafner, Timothy Lillicrap, Jimmy Ba, and Mohammad Norouzi.
\newblock Dream to control: Learning behaviors by latent imagination.
\newblock \emph{arXiv preprint arXiv:1912.01603}, 2019.

\bibitem[Yang et~al.(2023)Yang, Du, Ghasemipour, Tompson, Schuurmans, and Abbeel]{yang2023learning}
Mengjiao Yang, Yilun Du, Kamyar Ghasemipour, Jonathan Tompson, Dale Schuurmans, and Pieter Abbeel.
\newblock Learning interactive real-world simulators, 2023.

\bibitem[Latif et~al.(2023)Latif, Zaidi, Cuayahuitl, Shamshad, Shoukat, and Qadir]{latif2023transformers}
Siddique Latif, Aun Zaidi, Heriberto Cuayahuitl, Fahad Shamshad, Moazzam Shoukat, and Junaid Qadir.
\newblock Transformers in speech processing: A survey.
\newblock \emph{arXiv preprint arXiv:2303.11607}, 2023.

\bibitem[Rombach et~al.(2022)Rombach, Blattmann, Lorenz, Esser, and Ommer]{rombach2022high}
Robin Rombach, Andreas Blattmann, Dominik Lorenz, Patrick Esser, and Bj{\"o}rn Ommer.
\newblock High-resolution image synthesis with latent diffusion models.
\newblock In \emph{Proceedings of the IEEE/CVF conference on computer vision and pattern recognition}, pages 10684--10695, 2022.

\bibitem[Zou et~al.(2023)Zou, Kim, and Kang]{zou2023survey}
Hao Zou, Zae~Myung Kim, and Dongyeop Kang.
\newblock A survey of diffusion models in natural language processing.
\newblock \emph{arXiv preprint arXiv:2305.14671}, 2023.

\bibitem[Wei et~al.(2022)Wei, Wang, Schuurmans, Bosma, Xia, Chi, Le, Zhou, et~al.]{wei2022chain}
Jason Wei, Xuezhi Wang, Dale Schuurmans, Maarten Bosma, Fei Xia, Ed~Chi, Quoc~V Le, Denny Zhou, et~al.
\newblock Chain-of-thought prompting elicits reasoning in large language models.
\newblock \emph{Advances in Neural Information Processing Systems}, 35:\penalty0 24824--24837, 2022.

\bibitem[Goyal et~al.(2023)Goyal, Ji, Rawat, Menon, Kumar, and Nagarajan]{goyal2023think}
Sachin Goyal, Ziwei Ji, Ankit~Singh Rawat, Aditya~Krishna Menon, Sanjiv Kumar, and Vaishnavh Nagarajan.
\newblock Think before you speak: Training language models with pause tokens.
\newblock \emph{arXiv preprint arXiv:2310.02226}, 2023.

\bibitem[Hoover et~al.(2024)Hoover, Liang, Pham, Panda, Strobelt, Chau, Zaki, and Krotov]{hoover2024energy}
Benjamin Hoover, Yuchen Liang, Bao Pham, Rameswar Panda, Hendrik Strobelt, Duen~Horng Chau, Mohammed Zaki, and Dmitry Krotov.
\newblock Energy transformer.
\newblock \emph{Advances in Neural Information Processing Systems}, 36, 2024.

\bibitem[Wang et~al.(2022)Wang, Che, Li, Song, Pei, Bengio, and Li]{wang2022your}
Yezhen Wang, Tong Che, Bo~Li, Kaitao Song, Hengzhi Pei, Yoshua Bengio, and Dongsheng Li.
\newblock Your autoregressive generative model can be better if you treat it as an energy-based one.
\newblock \emph{arXiv preprint arXiv:2206.12840}, 2022.

\bibitem[Bhattacharyya et~al.(2020)Bhattacharyya, Rooshenas, Naskar, Sun, Iyyer, and McCallum]{bhattacharyya2020energy}
Sumanta Bhattacharyya, Amirmohammad Rooshenas, Subhajit Naskar, Simeng Sun, Mohit Iyyer, and Andrew McCallum.
\newblock Energy-based reranking: Improving neural machine translation using energy-based models.
\newblock \emph{arXiv preprint arXiv:2009.13267}, 2020.

\bibitem[Bakhtin et~al.(2021)Bakhtin, Deng, Gross, Ott, Ranzato, and Szlam]{bakhtin2021residual}
Anton Bakhtin, Yuntian Deng, Sam Gross, Myle Ott, Marc'Aurelio Ranzato, and Arthur Szlam.
\newblock Residual energy-based models for text.
\newblock \emph{Journal of Machine Learning Research}, 22\penalty0 (40):\penalty0 1--41, 2021.

\bibitem[Stojni{\'c} et~al.(2023)Stojni{\'c}, Gandhi, Yasuda, Lake, and Dillon]{stojnic2023commonsense}
Gala Stojni{\'c}, Kanishk Gandhi, Shannon Yasuda, Brenden~M Lake, and Moira~R Dillon.
\newblock Commonsense psychology in human infants and machines.
\newblock \emph{Cognition}, 235:\penalty0 105406, 2023.

\bibitem[Team(2023)]{Team_2023}
AlphaCode Team.
\newblock Alphacode 2 technical report.
\newblock December 2023.

\bibitem[Yao et~al.(2023)Yao, Yu, Zhao, Shafran, Griffiths, Cao, and Narasimhan]{yao2023tree}
Shunyu Yao, Dian Yu, Jeffrey Zhao, Izhak Shafran, Thomas~L. Griffiths, Yuan Cao, and Karthik Narasimhan.
\newblock Tree of thoughts: Deliberate problem solving with large language models, 2023.

\bibitem[Meta(2024{\natexlab{a}})]{Meta_2024}
Meta.
\newblock Introducing meta llama 3: The most capable openly available llm to date, April 2024{\natexlab{a}}.
\newblock URL \url{https://ai.meta.com/blog/meta-llama-3/}.

\bibitem[Evans(2003)]{evans2003two}
Jonathan St~BT Evans.
\newblock In two minds: dual-process accounts of reasoning.
\newblock \emph{Trends in cognitive sciences}, 7\penalty0 (10):\penalty0 454--459, 2003.

\bibitem[Alter et~al.(2007)Alter, Oppenheimer, Epley, and Eyre]{alter2007overcoming}
Adam~L Alter, Daniel~M Oppenheimer, Nicholas Epley, and Rebecca~N Eyre.
\newblock Overcoming intuition: metacognitive difficulty activates analytic reasoning.
\newblock \emph{Journal of experimental psychology: General}, 136\penalty0 (4):\penalty0 569, 2007.

\bibitem[Hinton(2002)]{hinton2002training}
Geoffrey~E Hinton.
\newblock Training products of experts by minimizing contrastive divergence.
\newblock \emph{Neural computation}, 14\penalty0 (8):\penalty0 1771--1800, 2002.

\bibitem[Du et~al.(2020{\natexlab{a}})Du, Li, Tenenbaum, and Mordatch]{du2020improved}
Yilun Du, Shuang Li, Joshua Tenenbaum, and Igor Mordatch.
\newblock Improved contrastive divergence training of energy based models.
\newblock \emph{arXiv preprint arXiv:2012.01316}, 2020{\natexlab{a}}.

\bibitem[Wang et~al.(2023)Wang, Wang, Liu, and Qiu]{wang2023energy}
Ze~Wang, Jiang Wang, Zicheng Liu, and Qiang Qiu.
\newblock Energy-inspired self-supervised pretraining for vision models.
\newblock \emph{arXiv preprint arXiv:2302.01384}, 2023.

\bibitem[Goyal et~al.(2017)Goyal, Ebrahimi~Kahou, Michalski, Materzynska, Westphal, Kim, Haenel, Fruend, Yianilos, Mueller-Freitag, et~al.]{goyal2017something}
Raghav Goyal, Samira Ebrahimi~Kahou, Vincent Michalski, Joanna Materzynska, Susanne Westphal, Heuna Kim, Valentin Haenel, Ingo Fruend, Peter Yianilos, Moritz Mueller-Freitag, et~al.
\newblock The" something something" video database for learning and evaluating visual common sense.
\newblock In \emph{Proceedings of the IEEE international conference on computer vision}, pages 5842--5850, 2017.

\bibitem[Kay et~al.(2017)Kay, Carreira, Simonyan, Zhang, Hillier, Vijayanarasimhan, Viola, Green, Back, Natsev, et~al.]{kay2017kinetics}
Will Kay, Joao Carreira, Karen Simonyan, Brian Zhang, Chloe Hillier, Sudheendra Vijayanarasimhan, Fabio Viola, Tim Green, Trevor Back, Paul Natsev, et~al.
\newblock The kinetics human action video dataset.
\newblock \emph{arXiv preprint arXiv:1705.06950}, 2017.

\bibitem[Computer(2023)]{together2023redpajama}
Together Computer.
\newblock Redpajama: an open dataset for training large language models, 2023.
\newblock URL \url{https://github.com/togethercomputer/RedPajama-Data}.

\bibitem[Zong et~al.(2023)Zong, Mac~Aodha, and Hospedales]{zong2023self}
Yongshuo Zong, Oisin Mac~Aodha, and Timothy Hospedales.
\newblock Self-supervised multimodal learning: A survey.
\newblock \emph{arXiv preprint arXiv:2304.01008}, 2023.

\bibitem[Yasar et~al.(2022)Yasar, Islam, and Iqbal]{yasar2022imprint}
Mohammad~Samin Yasar, Md~Mofijul Islam, and Tariq Iqbal.
\newblock Imprint: Interactional dynamics-aware motion prediction in teams using multimodal context.
\newblock \emph{ACM Transactions on Human-Robot Interaction}, 2022.

\bibitem[Islam et~al.(2022{\natexlab{b}})Islam, Yasar, and Iqbal]{islam2022maven}
Md~Mofijul Islam, Mohammad~Samin Yasar, and Tariq Iqbal.
\newblock Maven: A memory augmented recurrent approach for multimodal fusion.
\newblock \emph{IEEE Transactions on Multimedia}, 2022{\natexlab{b}}.

\bibitem[Team(2024)]{chameleonteam2024chameleon}
Chameleon Team.
\newblock Chameleon: Mixed-modal early-fusion foundation models, 2024.

\bibitem[Islam and Iqbal(2022)]{islam2022mumu}
Md~Mofijul Islam and Tariq Iqbal.
\newblock Mumu: Cooperative multitask learning-based guided multimodal fusion.
\newblock In \emph{Proceedings of the AAAI conference on artificial intelligence}, volume~36, pages 1043--1051, 2022.

\bibitem[Islam and Iqbal(2020)]{islam2020hamlet}
Md~Mofijul Islam and Tariq Iqbal.
\newblock Hamlet: A hierarchical multimodal attention-based human activity recognition algorithm.
\newblock In \emph{2020 IEEE/RSJ International Conference on Intelligent Robots and Systems (IROS)}, pages 10285--10292. IEEE, 2020.

\bibitem[Islam and Iqbal(2021)]{islam2021multi}
Md~Mofijul Islam and Tariq Iqbal.
\newblock Multi-gat: A graphical attention-based hierarchical multimodal representation learning approach for human activity recognition.
\newblock \emph{IEEE Robotics and Automation Letters}, 6\penalty0 (2):\penalty0 1729--1736, 2021.

\bibitem[Liu et~al.(2024{\natexlab{b}})Liu, Li, Wu, and Lee]{liu2024visual}
Haotian Liu, Chunyuan Li, Qingyang Wu, and Yong~Jae Lee.
\newblock Visual instruction tuning.
\newblock \emph{Advances in neural information processing systems}, 36, 2024{\natexlab{b}}.

\bibitem[Berglund et~al.(2023)Berglund, Tong, Kaufmann, Balesni, Stickland, Korbak, and Evans]{berglund2023reversal}
Lukas Berglund, Meg Tong, Max Kaufmann, Mikita Balesni, Asa~Cooper Stickland, Tomasz Korbak, and Owain Evans.
\newblock The reversal curse: Llms trained on" a is b" fail to learn" b is a".
\newblock \emph{arXiv preprint arXiv:2309.12288}, 2023.

\bibitem[Patel(2023)]{Patel_2023}
Dwarkesh Patel.
\newblock Will scaling work?, December 2023.
\newblock URL \url{https://www.dwarkeshpatel.com/p/will-scaling-work}.

\bibitem[Villalobos et~al.(2022)Villalobos, Sevilla, Heim, Besiroglu, Hobbhahn, and Ho]{villalobos2022will}
Pablo Villalobos, Jaime Sevilla, Lennart Heim, Tamay Besiroglu, Marius Hobbhahn, and Anson Ho.
\newblock Will we run out of data? an analysis of the limits of scaling datasets in machine learning.
\newblock \emph{arXiv preprint arXiv:2211.04325}, 2022.

\bibitem[Gunasekar et~al.(2023)Gunasekar, Zhang, Aneja, Mendes, Del~Giorno, Gopi, Javaheripi, Kauffmann, de~Rosa, Saarikivi, et~al.]{gunasekar2023textbooks}
Suriya Gunasekar, Yi~Zhang, Jyoti Aneja, Caio C{\'e}sar~Teodoro Mendes, Allie Del~Giorno, Sivakanth Gopi, Mojan Javaheripi, Piero Kauffmann, Gustavo de~Rosa, Olli Saarikivi, et~al.
\newblock Textbooks are all you need.
\newblock \emph{arXiv preprint arXiv:2306.11644}, 2023.

\bibitem[Meta(2024{\natexlab{b}})]{Meta_2024b}
Meta.
\newblock Our responsible approach to meta ai and meta llama 3, April 2024{\natexlab{b}}.
\newblock URL \url{https://ai.meta.com/blog/meta-llama-3-meta-ai-responsibility/}.

\bibitem[Islam et~al.(2022{\natexlab{c}})Islam, Mirzaiee, Gladstone, Green, and Iqbal]{islam2022caesar}
Md~Mofijul Islam, Reza Mirzaiee, Alexi Gladstone, Haley Green, and Tariq Iqbal.
\newblock Caesar: An embodied simulator for generating multimodal referring expression datasets.
\newblock \emph{Advances in Neural Information Processing Systems}, 35:\penalty0 21001--21015, 2022{\natexlab{c}}.

\bibitem[Islam et~al.(2023{\natexlab{b}})Islam, Gladstone, and Iqbal]{patron}
Md~Mofijul Islam, Alexi Gladstone, and Tariq Iqbal.
\newblock Patron: perspective-aware multitask model for referring expression grounding using embodied multimodal cues.
\newblock In \emph{Proceedings of the Thirty-Seventh AAAI Conference on Artificial Intelligence and Thirty-Fifth Conference on Innovative Applications of Artificial Intelligence and Thirteenth Symposium on Educational Advances in Artificial Intelligence}, AAAI'23/IAAI'23/EAAI'23. AAAI Press, 2023{\natexlab{b}}.
\newblock ISBN 978-1-57735-880-0.
\newblock \doi{10.1609/aaai.v37i1.25177}.
\newblock URL \url{https://doi.org/10.1609/aaai.v37i1.25177}.

\bibitem[Trinh et~al.(2024)Trinh, Wu, Le, He, and Luong]{trinh2024solving}
Trieu~H Trinh, Yuhuai Wu, Quoc~V Le, He~He, and Thang Luong.
\newblock Solving olympiad geometry without human demonstrations.
\newblock \emph{Nature}, 625\penalty0 (7995):\penalty0 476--482, 2024.

\bibitem[Lee et~al.(2024)Lee, Wattanawong, Kim, Mangalam, Shen, Anumanchipali, Mahoney, Keutzer, and Gholami]{lee2024llm2llm}
Nicholas Lee, Thanakul Wattanawong, Sehoon Kim, Karttikeya Mangalam, Sheng Shen, Gopala Anumanchipali, Michael~W Mahoney, Kurt Keutzer, and Amir Gholami.
\newblock Llm2llm: Boosting llms with novel iterative data enhancement.
\newblock \emph{arXiv preprint arXiv:2403.15042}, 2024.

\bibitem[LeCun et~al.(2006)LeCun, Chopra, Hadsell, Ranzato, and Huang]{lecun2006tutorial}
Yann LeCun, Sumit Chopra, Raia Hadsell, M~Ranzato, and Fujie Huang.
\newblock A tutorial on energy-based learning.
\newblock \emph{Predicting structured data}, 1\penalty0 (0), 2006.

\bibitem[Song and Kingma(2021)]{song2021train}
Yang Song and Diederik~P Kingma.
\newblock How to train your energy-based models.
\newblock \emph{arXiv preprint arXiv:2101.03288}, 2021.

\bibitem[Du et~al.(2021)Du, Li, Sharma, Tenenbaum, and Mordatch]{du2021unsupervised}
Yilun Du, Shuang Li, Yash Sharma, Josh Tenenbaum, and Igor Mordatch.
\newblock Unsupervised learning of compositional energy concepts.
\newblock \emph{Advances in Neural Information Processing Systems}, 34:\penalty0 15608--15620, 2021.

\bibitem[Du et~al.(2020{\natexlab{b}})Du, Li, and Mordatch]{du2020compositional}
Yilun Du, Shuang Li, and Igor Mordatch.
\newblock Compositional visual generation with energy based models.
\newblock \emph{Advances in Neural Information Processing Systems}, 33:\penalty0 6637--6647, 2020{\natexlab{b}}.

\bibitem[Black et~al.(2022)Black, Biderman, Hallahan, Anthony, Gao, Golding, He, Leahy, McDonell, Phang, Pieler, Prashanth, Purohit, Reynolds, Tow, Wang, and Weinbach]{black2022gptneox20b}
Sid Black, Stella Biderman, Eric Hallahan, Quentin Anthony, Leo Gao, Laurence Golding, Horace He, Connor Leahy, Kyle McDonell, Jason Phang, Michael Pieler, USVSN~Sai Prashanth, Shivanshu Purohit, Laria Reynolds, Jonathan Tow, Ben Wang, and Samuel Weinbach.
\newblock Gpt-neox-20b: An open-source autoregressive language model, 2022.

\bibitem[Falcon(2019)]{falcon2019pytorch}
William~A Falcon.
\newblock Pytorch lightning.
\newblock \emph{GitHub}, 3, 2019.

\end{thebibliography}
}

\appendix
\clearpage
\section{Additional Experimentation: {\pa} for Natural Language Processing}
\label{sec:nlp_exp}
\begin{figure}[h]
    \begin{center}
    \scriptsize
    \begin{tabular}{ccc}
        \hspace{-.3cm}\includegraphics[width=0.33\columnwidth]{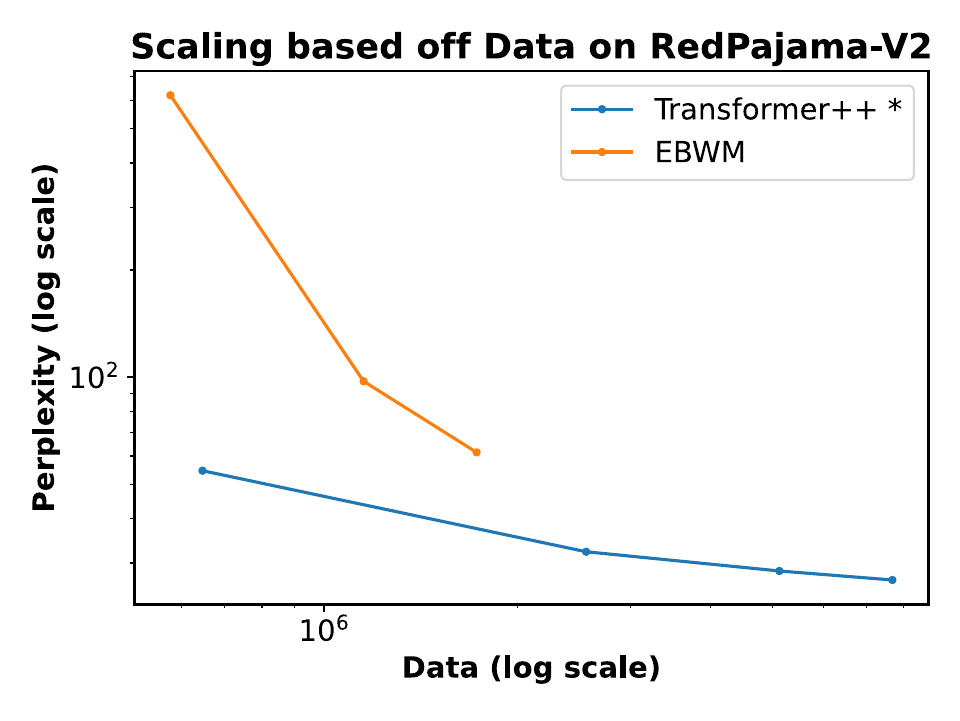} &
        \hspace{-.3cm}\includegraphics[width=0.34\columnwidth]{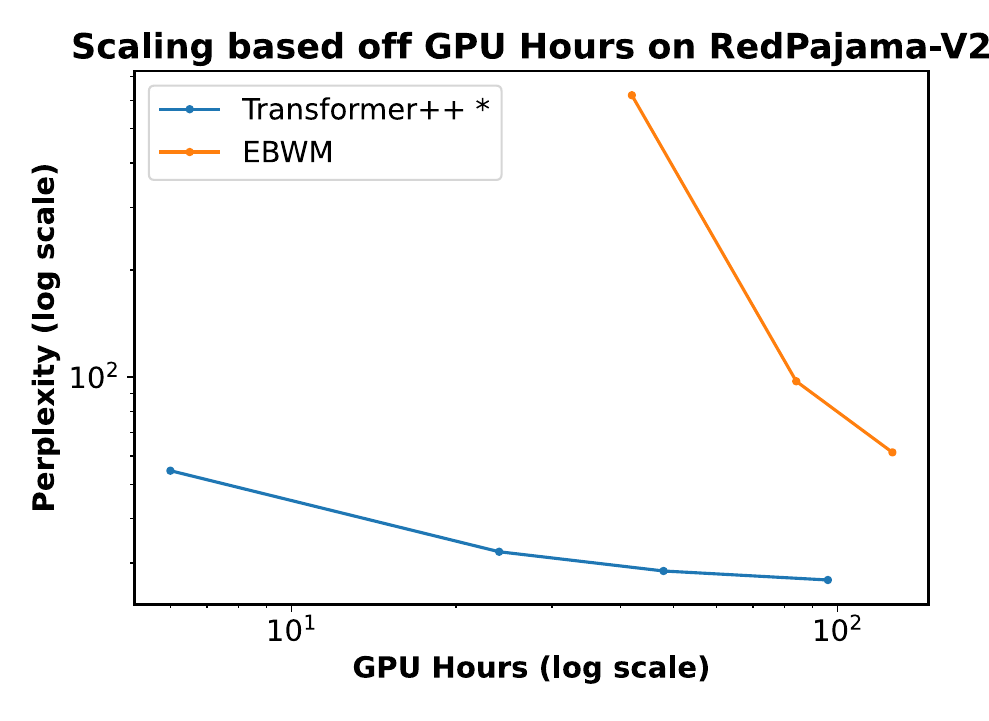} &
        \hspace{-.25cm}\includegraphics[width=0.33\columnwidth]{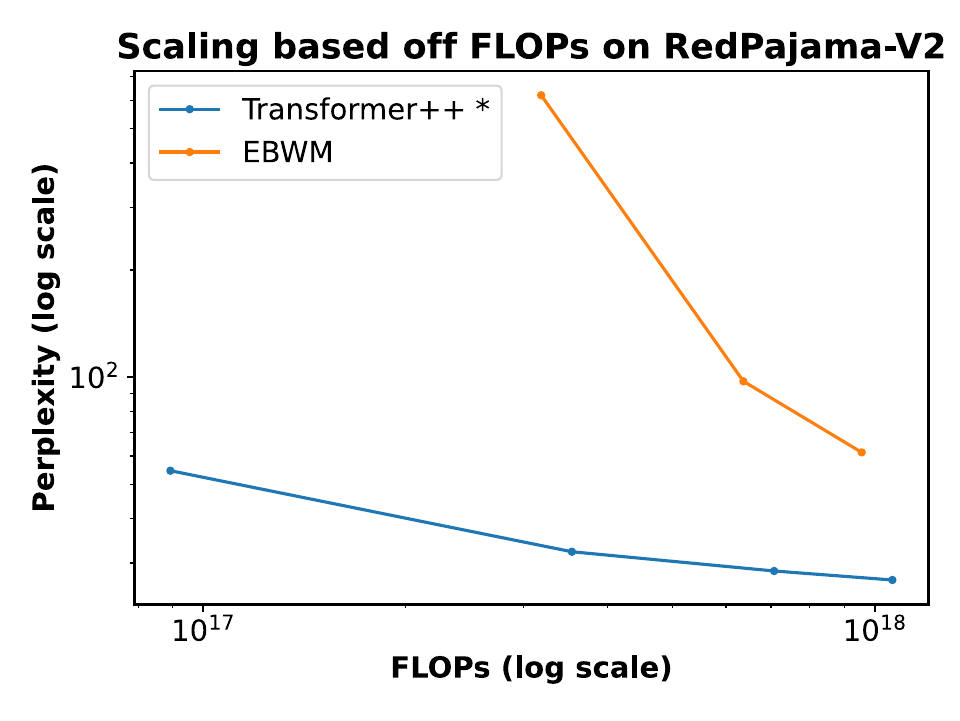} \\
        (a) Scaling for data across epochs. &
        (b) Scaling for A100 GPU Hours. &
        (c) Scaling for FLOPs. \\
    \end{tabular}

    \caption{The perplexity across different scales of data, GPU hours, and FLOPs (lower and farther to the left curves are better) for the RedPajama-V2 dataset. The experimental results show the promising scaling of NLP models using EBWM. We believe that with more resources to search over different hyperparameters {\pa} could perform significantly better. {\pa} used a smaller batch size due to involving more memory, and hence was much slower, resulting in it having less datapoints given our compute limit. *The first two points for the Transformer++ were perplexity values on the training set.}
    
    \label{fig:nlp_scaling}
    \end{center}
\end{figure}

For all experiments in Natural Language Processing (NLP), we utilize the RedPajama-V2 dataset \cite{together2023redpajama} $100B$ sample \href{https://huggingface.co/datasets/togethercomputer/RedPajama-Data-V2}{from HuggingFace}, with a manually created train and validation split of $65,818,073$ and $330,745$ respectively. The training objective is the traditional language modeling objective of predicting the next token in a sequence. We report scaling for data, GPU hours, and FLOPs of {\pa} compared to {\baseline} in Fig.~\ref{fig:nlp_scaling}. Despite not being able to tune hyperparameters due to limited computational resources, the results demonstrate the scalability of {\pa} in data efficiency.

\section{Future Works and Broader Impact }
{\pa}, being qualitatively different from existing autoregressive architectures, while having all four cognitive facets described, opens several future research directions.

\subsection{{\pa} for other domains}
We designed {\parch} as a modality agnostic autoregressive architecture. Consequently, we find it highly plausible {\pa} could be used in domains other than NLP and CV. For example, sequential data in Audio Processing, Graph based learning, Time Series Analysis, and Tactile \& Haptic Feedback, could be modeled similarly to CV/NLP data with {\pa}.

\subsection{Multimodal {\pa}}
Multimodal learning has progressed rapidly within recent years \cite{zong2023self, yasar2022imprint, islam2022maven, chameleonteam2024chameleon, islam2022mumu, islam2020hamlet, islam2021multi, liu2024visual}. {\pa}, being qualitatively different from existing architectures, offers an exciting future research direction for extensions to multiple modalities.

\subsection{Reversal curse}
Recently, a phenomenon known as ``The Reversal Curse'' has been observed where LLMs fail to learn a symmetric mapping \cite{berglund2023reversal} ``B is A'' despite learning ``A is B''. For example, LLMs trained on an example such as ``Q: Who is Tom Cruise’s mother? A: Mary Lee Pfeiffer'' often fail to generalize to know the answer to the reverse question of ``Who is Mary Lee Pfeiffer’s son?'' Remarkably, the Reversal Curse has manifested itself in LLMs regardless of the size or scale \cite{berglund2023reversal}---probing researchers to investigate whether there are fundamental limitations to {\baseline}. One predicted cause of The Reversal Curse is the nature of gradient updates being passed only to tokens within context. That is, while learning the mapping ``A is B'', none of B's tokens are within context meaning they do not receive gradient updates. In this paper, we hypothesize using {\pa} would help reduce this phenomenon, as both A's and B's tokens are within context during gradient updates due to future state predictions being made in the input space. Therefore, an exciting research direction would be investigating whether this hypothesis is correct in LLMs trained with {\pa}.

\subsection{World models for acting}
In this work, we focus on a specific instance of world models where a default ``no-op'' action is taken. However, more complex world models often consist of forward rollouts conditioned off of specific actions being taken. In this paradigm, {\pa} offers high promise due to the nature of EBMs. Particularly, given a model trained to estimate the unnormalized joint distribution of the current context, future, as well as the next action, such world models could implicitly be used as policies to generate actions to achieve a specific state. This would involve holding the current context constant and minimizing the energy by passing the gradient back to the action inputs and future state predictions. Thus, world models trained in this manner become capable of more than just predicting the future, but also in decision making to achieve a specific goal state. 

\subsection{Improving traditional autoregressive model predictions}
\label{sec:complementary}
As demonstrated in \cite{bakhtin2021residual, bhattacharyya2020energy}, EBMs can be used to improve the quality of generated text from language models. In the context of {\pa}, this is possible for world models across domains. This process would involve making a forward pass with a traditional autoregressive world model, passing this model's output into {\pa}, and then using MCMC to further improve upon the {\baseline} prediction. This combination allows for fast inference when necessary, such as when low latency is a priority. It also allows for the cognitive facets described in Section \ref{sec:cognitive_facets}, potentially enabling for a situation where different more challenging problems get the extra computational resources needed to solve them. This is one of the reasons we consider {\pa} \textit{complementary} to {\baseline}, rather than being a competitor.

\subsection{Data Augmentation}
\label{sec:data_aug}

One problem towards further scaling of foundation models is the scale of data \cite{Patel_2023, villalobos2022will}, as, especially in the realm of NLP, much of the data on the open internet has been exhausted. As a consequence the creation of synthetic datasets has become common \cite{gunasekar2023textbooks, Meta_2024b, islam2022caesar, patron}. With {\baseline}, the process of data augmentation usually involves taking advantage of algorithmically verifiable text such as code or math \cite{trinh2024solving} or synonym replacement \cite{lee2024llm2llm} to generate plausible next tokens. With {\pa}, the process of data augmentation is significantly easier due to the approximation of a joint distribution \(p(x_1, ..., x_t, x_{t+1)}\) rather than the conditional distribution \(p(x_{t+1} | x_1, ..., x_t)\) {\baseline} approximate. This was demonstrated in Section \ref{sec:mcmc}, where a different number of MCMC steps were used, as each of these MCMC steps can be seen as a different data sample (\(p(x_1, ..., x_t, \hat{x}_{t+1)}\)) due to there being a different condition (\(\hat{x}_{t+1)}\)). A promising future direction is an investigation of the extent to which data augmentation with {\pa} can be done without reducing performance, and the benefits this can have in reducing the amount of pre-augmentation data needed.

\subsection{Intelligent Search}
\label{sec:search}
Due to a lack of computational resources, we were unable to train models in the regime of \(1,000\) or more A100 GPU Hours. Therefore, the usage of {\pa} for intelligent search remains untested at scale. We leave it to future work to further scale with more GPUs and investigate the qualitative differences in searching the state space with {\baseline} versus with {\pa} as described in Section \ref{sec:intuition}.

\subsection{Societal impact}
The ability to achieve human level cognition with AI offers benefits in multiple domains. As such, {\pa} offers several potential positive impacts, through the enabling of AI to be more similar to human cognition. On the other hand, it's also possible that more intelligent AI models trained using {\pa} could be misused for harm by malicious actors.

\section{Ablation Studies}

\subsection{Losses}
\label{sec:supp_losses}
In addition to EBM training over sequences using a reconstruction loss \cite{wang2023energy}, we develop a new approach involving the prediction of ground truth energy labels--we call this {\ebma}. This allows us to model energy-based model training as a regression task:

\begin{align}
\mathcal{L}_e = D(E, \hat{E})
\end{align}

Where $\mathcal{L}_e$ denotes the energy loss. To generate these ground truth energy labels, we use the distance between the embeddings/features of a ground truth and predicted future state. 

\begin{align}
E = D(\mathbf{z}, \hat{\mathbf{z}}).
\end{align}

As such, we call this {\el}. 

One benefit of using encoder representations for labeling the ground truth energy value of a given future prediction is the ability to regularize the output energy space. Particularly, since we use the cosine similarity as a distance function, labeled energy values by default are within the range $[-1,  1]$. In this work, we further scale this range to be $[0,  1]$, with lower energy values indicating higher compatibility future state predictions.

Utilizing this characteristic, we add an additional regularization term, which we call the Out-of-Bounds loss. This regularization term, in addition to the energy loss term, enforces predicted energies to be within a specified range, in this case $[0,  1]$:

$$
\mathcal{L}_b = \max(0, \hat{E} - 1) + \max(0, -\hat{E})
$$

The results for using these two losses are shown in Table \ref{tab:design_choices}. Overall, we find these losses are \textbf{not} helpful in increasing stability, and that the rather simple approach of a single reconstruction objective works best.

\subsection{Condition for MCMC}
\begin{figure}[ht]
    \centering
    \begin{tabular}{cc} 
        \includegraphics[width=0.45\columnwidth]{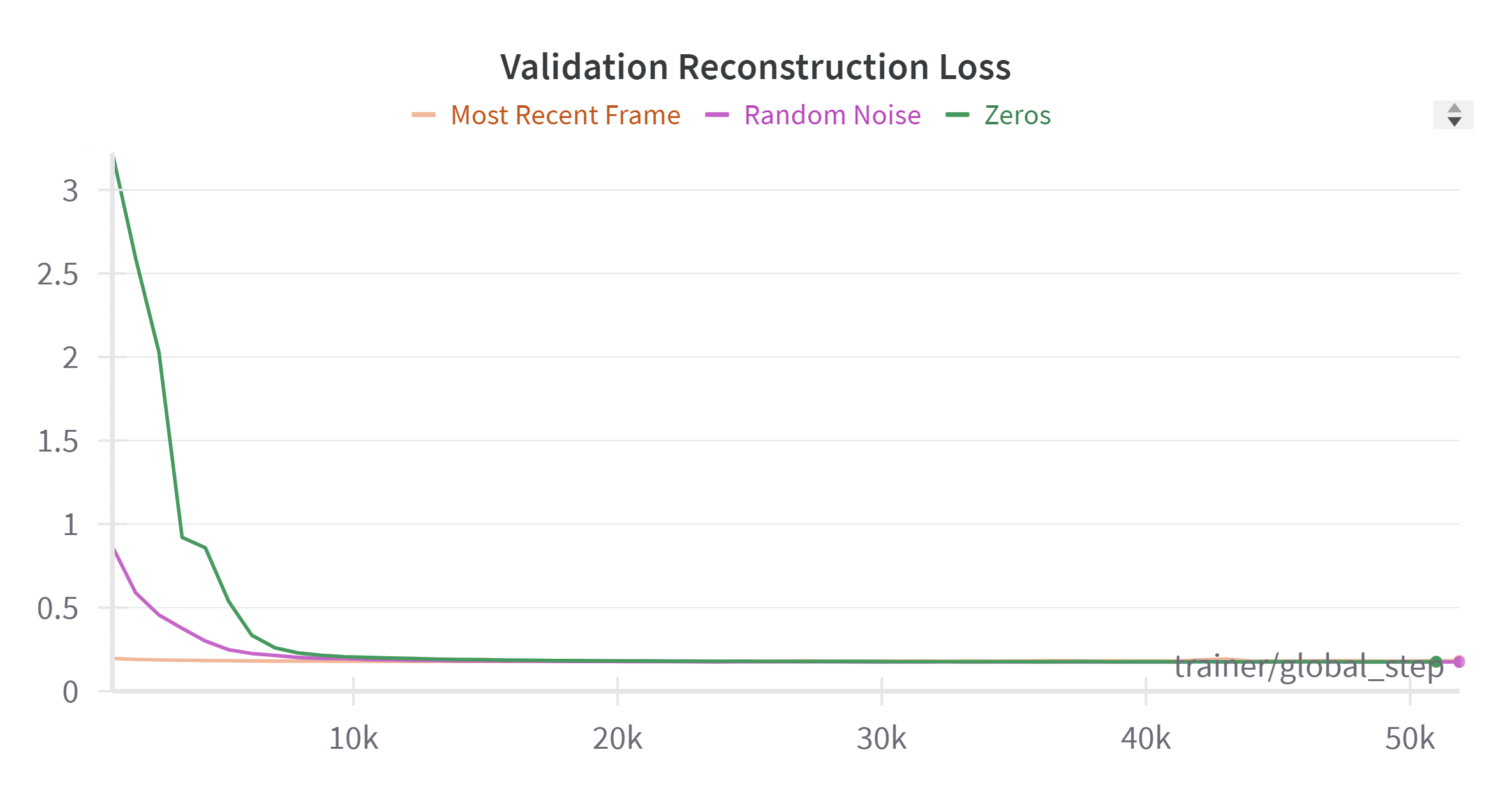} &
        \includegraphics[width=0.45\columnwidth]{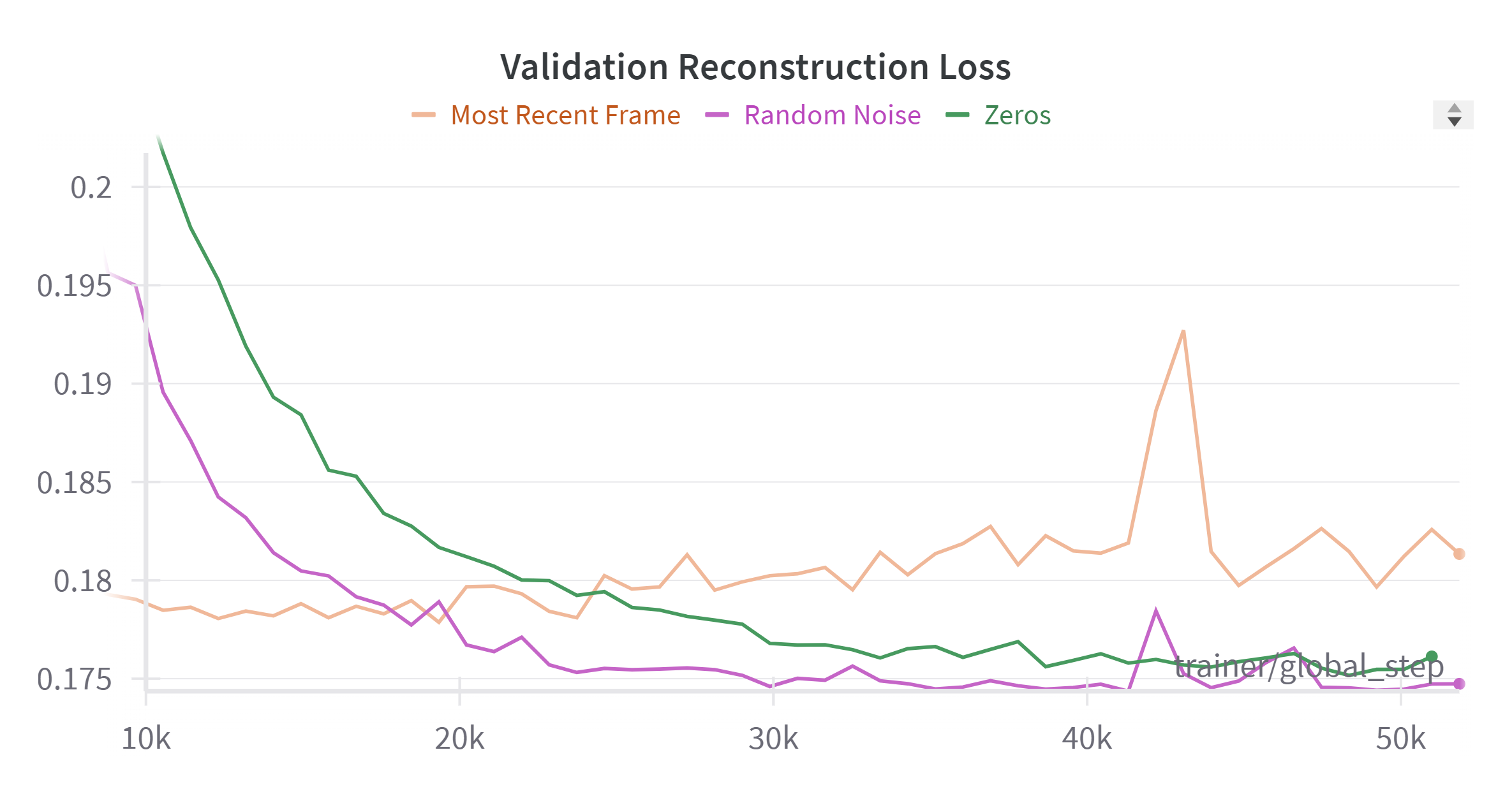} \\
        (a) Earlier validation scaling. &
        (b) Later validation scaling. \\
    \end{tabular}
    \caption{Validation loss curves with different MCMC conditions. Although the loss is initially better when conditioning on the most recent frame due to a high temporal overlap in videos, conditioning on random noise ultimately achieves the best validation loss. The validation loss of conditioning on the most recent frame starts at $0.196$.}
    \label{fig:mcmc_conditions}
\end{figure}

Energy Based Models are a family of generative models that predict the unnormalized density, or compatibility, of a configuration of variables. Through the training a robust energy space, inference from Energy Based Models involves traversal of this energy space through gradient descent. Consequentially, one of the most common sampling techniques for Energy Based Models is Markov Chain Monte Carlo (MCMC). At each time step of the Markov Chain, the current predicted $\hat{y}$ is inputted into the model, an energy is computed, and through gradient descent this energy function is descended. Historically MCMC has begun from random noise \cite{lecun2006tutorial, song2021train}, resulting in a very long Markov Chain to reach convergence. However, recently works have experimented with MCMC from a condition \cite{du2021unsupervised, du2020compositional, wang2023energy}. This has the advantage of reducing the number of steps to convergence within the energy function. We experimented with a couple of different approaches for our MCMC condition in CV including conditioning on the most recent recent frame, conditioning on all zeros, and conditioning from random noise. We found that although conditioning on the most recent frame achieves better initial convergence due to the high temporal consistency in videos, random noise scaled better. Plots for losses based on different conditions are shown in Fig.~\ref{fig:mcmc_conditions}.

\section{Additional {\pa} Details}
\label{sec:training_details}


\begin{table*}[!t]
\centering
\caption{\textbf{Hyperparameters for \textbf{\baseline}.} Model information and hyperparameters for our {\baseline} experiments.}
\begin{tabular}{lcc}
\toprule
\textbf{Hyperparameter} & \textbf{CV} & \textbf{NLP} \\
\toprule
Batch Size per GPU & $64$ & $128$ \\
Effective Batch Size & $192$ & $384$ \\
Epochs & \multicolumn{2}{c}{$400$} \\
Optimizer & \multicolumn{2}{c}{AdamW} \\
Optimizer Momentum & \multicolumn{2}{c}{$\beta_1,\beta_2=0.9,0.999$} \\
Freeze Encoder Epochs & \multicolumn{2}{c}{$2000$} \\
Learning Rate (LR) & $1e-3$ & $6e-4$ \\
LR Schedule & \multicolumn{2}{c}{Linear warm up cosine decay} \\
Warmup steps & \multicolumn{2}{c}{$1e4$} \\
Warmup Base LR Divider & \multicolumn{2}{c}{$20$} \\
Minimum LR Scale & \multicolumn{2}{c}{$10$} \\
Gradient Clip Value & \multicolumn{2}{c}{$1$} \\
Transformer Blocks & \multicolumn{2}{c}{$12$} \\
Multi-headed Attention Heads & \multicolumn{2}{c}{$12$} \\
Weight Decay & \multicolumn{2}{c}{$0.01$} \\
Context Length & $16$ & $256$ \\
ViT Backbone Model & DINOv2~\cite{oquab2023dinov2} & - \\
ViT Backbone Size & Base & - \\
Image Dimension & $224$x$224$ & - \\
Tokenizer & - & EleutherAI/gpt-neox-20b~\cite{black2022gptneox20b} \\
Vocab Size & - & $50277$ \\
\bottomrule
\end{tabular}
\label{tab:hparams_baseline}
\end{table*}


\begin{table*}[!t]
\centering
\caption{\textbf{Hyperparameters for \textbf{\pa} experiments.} Model information and hyperparameters for our {\pa} experiments.}
\begin{tabular}{lcc}
\toprule
\textbf{Hyperparameter} & \textbf{CV} & \textbf{NLP} \\
\toprule
Batch Size per GPU & $64$ & $24$ \\
Effective Batch Size & $192$ & $72$ \\
Epochs & \multicolumn{2}{c}{$400$} \\
Optimizer & \multicolumn{2}{c}{AdamW} \\
Optimizer Momentum & \multicolumn{2}{c}{$\beta_1,\beta_2=0.9,0.999$} \\
Learning Rate (LR) & \multicolumn{2}{c}{$2e-4$} \\
LR Schedule & \multicolumn{2}{c}{Linear warm up cosine decay} \\
Warmup Steps & \multicolumn{2}{c}{$1e4$} \\
Warmup Base LR Divider & \multicolumn{2}{c}{$20$} \\
Minimum LR Scale & \multicolumn{2}{c}{$10$} \\

Energy Loss Coefficient & \multicolumn{2}{c}{$0$} \\
Out-of-bounds Loss Coefficient & \multicolumn{2}{c}{$0$} \\
Reconstruction Coefficient & \multicolumn{2}{c}{$60$} \\

MCMC Steps & $4$ & $2$ \\
MCMC Step Size & $3e4$ & $3e5$ \\
Learnable MCMC Step Size & \multicolumn{2}{c}{\cmark} \\
MCMC Step Size LR Multiplier & $2e5$ & $2e6$ \\
Clamp Future Gradient & \multicolumn{2}{c}{\cmark} \\
Langevin Dynamics Noise & \multicolumn{2}{c}{$0$}\\

Weight Decay & \multicolumn{2}{c}{$0.01$} \\
Context Length & $16$ & $256$ \\
Transformer Blocks & \multicolumn{2}{c}{$12$} \\
Multi-headed Attention Heads & \multicolumn{2}{c}{$12$} \\
Gradient Clip Value & \multicolumn{2}{c}{$1$} \\
End MLP Layers & \multicolumn{2}{c}{$1$} \\

ViT Backbone Model & DINOv2~\cite{oquab2023dinov2} & - \\
ViT Backbone Size & Base & - \\
Image Dimension & $224$x$224$ & - \\
Tokenizer & - & EleutherAI/gpt-neox-20b~\cite{black2022gptneox20b} \\
Vocab Size & - & $50277$ \\

\bottomrule
\end{tabular}
\label{tab:hparams_pfc}
\end{table*}

Tables~\ref{tab:hparams_baseline} and~\ref{tab:hparams_pfc} specify model information and hyperparameters for each experiment.

The effective batch size for each training run is triple the batch size per GPU, as we train with three Nvidia A100 GPUs for all experiments. Note that although we specify for the models to train for $400$ epochs, this was not completed due to having a limited compute budget. We use zero dropout across all training, and regularize model weights by using weight decay. To schedule the learning rate, we use a linear warm up with cosine annealing. We warm up for $1e4$ steps and decay by at most $10x$ the base learning rate. We set the coefficients for energy loss and out-of-bounds loss to zero, due to the results found in Table \ref{tab:design_choices}, removing them from the loss calculation.

We utilized the Llama 2 transformer implementation \cite{touvron2023llama} for {\baseline} and used this implementation as the backbone upon which we built {\parch}. We use the following rule for scaling all learning rates (all learning rates shown in the table were base learning rates that got scaled according to this rule): \\

\[
\text{lr} = \text{base\_learning\_rate} * \text{effective\_batch\_size} / 256
\]

Where the \(\text{effective\_batch\_size}\) is calculated based off the batch size per GPU and the number of GPUs multiplied (since we use DDP).

\subsection{Reproducibility}
We utilize a singularity container and PyTorch Lightning \cite{falcon2019pytorch} for all experiments. We seed all libraries using PyTorch Lightning with a seed of $33$. We plan to release all code, containers for execution, as well as a comprehensive setup guide in the future. All scaling experiments conducted within the main paper were done with three Nvidia A100 $80$ GB GPU's for at most $72$ hours (for an effective maximum amount of $216$ GPU hours). We plan to release all source code on GitHub publicly soon.

\subsection{Improving Stability}
We experiment with several different hyperparameters to increase the stability of the {\pa} approach.
\subsubsection{Clamping Gradients}
We clamp the future gradient for MCMC to help reconstruct images and avoid exploding gradient on backward passes. As clamping gradients for regular backpropagation is common to prevent loss spikes, this was an intuitive and helpful inductive bias for improving stability. The results in Table \ref{tab:design_choices} demonstrate that this choice improves stability.

\subsubsection{High MCMC step size}
In addition to setting a high MCMC step size, we make the MCMC step size a learnable parameter. We calculate its learning rate by multiplying the model's learning rate by the MCMC step size learning rate multiplier. We find that the initial values for the MCMC step size have a large effect on the magnitude of gradients generated, and hence the stability of the model. Particularly, a smaller MCMC step size resulted in larger generated gradients. The results for a lower MCMC step size in Table \ref{tab:design_choices} demonstrate this effect, with a smaller MCMC step size being less stable.

\subsection{More intuition}

\textbf{Future state prediction in the input versus output space:}
We posit that there exists a fundamental distinction between the internal representations associated with model inputs and those concerning model outputs. Specifically, models generate internal representations \textit{of} inputs, as these serve as the foundational elements that exclusively dictate the model's behavior at any given point in time. Conversely, since outputs do not affect a model's behavior at a given point in time, we contend that models construct representations \textit{for predicting} such outputs. This distinction leads us to a consequential insight: models may not achieve a genuine understanding of outputs in the same way they understand inputs. Instead, they primarily excel in making predictions based on these output-focused representations. This foundational difference implies that while models may develop an intricate understanding of context when presented as input data, such understanding does not naturally extend to the realm of predictions made in the output space---indicating a limitation in models' abilities to comprehend future state predictions in the output space. This intuition further supports the principles behind {\pa}.

\subsection{Energy-Based Transformer Full Implementation}
\label{sec:ebt_full}
As described in Section \ref{sec:ebt_intro}, {\parch} involves two separate tensors--one for past states and one for predicted future states. We denote these as \(z_1^n\) and \(\hat{z}_1^n\) where \(z\) are known past states and \(\hat{z}\) are predicted future states. The intended attention scores matrix is the following:

\[
\text{scores} = 
\begin{bmatrix}
\alpha_{z_1,z_1} & \alpha_{z_1,\hat{z}_2} & 0 & \ldots & 0 \\
\alpha_{z_2,z_1} & \alpha_{z_2,z_2} & \alpha_{z_2,\hat{z}_3} & \ldots & 0 \\
\ldots & \ldots & \ldots & \ldots & \ldots \\
\alpha_{z_n,z_1} & \alpha_{z_n,z_2} & \alpha_{z_3,z_3} & \ldots & \alpha_{z_n,\hat{z}_{n+1}} \\
\end{bmatrix}.
\]

To compute Equation \ref{math:scores_ebm}, we first need to append a column to the right side of the \text{unnormalized\_scores\_p} matrix, as the size of the matrix is currently $n-1 \times n-1$, but we need to have $n$ representations within context. After doing this, we first mask out the superdiagonal, to ensure that the probabilities in the score matrix only correspond to the values of the predicted future states with itself. This masking operation is done through elementwise multiplication of a matrix with 1's everywhere except the superdiagonal, which has 0's. Then, we compute the self-attention scores of each predicted future state with itself, using the following equation: \\

\begin{equation}
\text{z\_p\_self\_attention} = sum(Q_p * K_p),
\end{equation}

where the $*$ indicates the Hadamard product and the sum is across the fourth, attention head, dimension. Using a superdiagonal mask again, we set the diagonal of the \text{unnormalized\_scores\_p} to these values. Now, after applying the softmax: \\

\begin{equation}
    \text{scores\_p} = \text{softmax}\left( \text{unnormalized\_scores\_p} \right),
\end{equation}

we have the intended scores matrix shown in Equation \ref{math:scores_ebm}. However, one more barrier towards finally extracting all updated \(z_{p 1}^n\) representations is the fact that we cannot simply multiply this resulting scores matrix by the values matrix, as each element of the superdiagonal corresponds to a different predicted next future state. Thus, using similar techniques to before, we first clone and then extract the superdiagonal from this scores matrix using a diagonal mask.

After extracting the superdiagonal, we can multiply the resulting scores matrix by the $V_o$ matrix to get all of the representations summed together of each predicted future state with all past states. This is represented as the following matrix multiplication: \\

\begin{equation}
z_{p 1}^n = \text{scores\_p} \cdot V_o.
\end{equation}

As we also need to add the representation of each predicted future state weighted with its own attention score (what was extracted on the superdiagonal), we perform another Hadamard product of the \(V_p\) matrix with the cloned superdiagonal to get these values, and then add these element wise to the \(z_{p 1}^n\) representations. Now, we have computed the intended representations involving the scores matrix shown in Equation \ref{math:scores_ebm}. Thus, \(z_1^n\) and \(\hat{z}_1^n\) are updated using \(z_{o 1}^n\) and \(z_{p 1}^n\) respectively, by multiplying these tensors by the output weight matrix $W_o$ .
\section{Counterarguments}
\label{sec:counterarguments}

\subsection{System 2 thinking}
In this paper, one of the major claims was that {\baseline} cannot do System 2 thinking like humans involving leveraging a dynamic amount of computation for predictions. However, there are common counterarguments to this, which we address, in hopes of clarifying why we believe this is not currently possible.

\subsubsection{Chain-of-thought reasoning}
First, chain-of-thought involves reasoning over a discrete state space, which limits the granularity of ``thoughts''. Second, chain-of-thought is not a capability internal to a model architecture, but rather something done externally over tokens. This capability should be built into an architecture and trained. Third, each individual token has a finite amount of computation to produce it, meaning that models cannot reason over each individual token used. Ideally, just as how when humans think ``step by step,'' each step takes varying amounts of times, models could focus on each token for a specific amount of time until deemed adequate (as done in {\pa}).

\subsubsection{Pause token}
Recently, seeking to achieve human-like reasoning capabilities researchers investigated the usage of a ``pause token'' \cite{goyal2023think}. Training models in this manner achieves better approximation towards cognitive facet (3) dynamic computation allocation \ref{item:dynamic_computation}. However, this approach is ultimately limited for the same reason that diffusion models are, in that they have no way of determining when a sufficient amount of computation has been used. This requires the ability to discriminate the plausibility of predictions, as described in Section \ref{item:prediction_evaluation}.


\end{document}